\documentclass[11pt]{article}

\usepackage[preprint]{acl}

\usepackage{times}
\usepackage{adjustbox}
\usepackage{latexsym}
\usepackage{amsmath}

\usepackage{amssymb}
\usepackage{amsfonts}
\usepackage{hyperref}
\usepackage{subcaption}
\usepackage{tcolorbox}
\usepackage{xeCJK}

\usepackage{booktabs}
\usepackage{pifont}
\usepackage{xcolor}
\usepackage{array}
\usepackage{multirow}
\usepackage{makecell}

\usepackage{arydshln}

\newcommand{\cmark}{\textcolor{green!60!black}{\ding{51}}}
\newcommand{\xmark}{\textcolor{red!70!black}{\ding{55}}}
\newcommand{\pmark}{\textcolor{orange!80!black}{\ding{108}}}

\usepackage{gradient-text}

\usepackage[english,bidi=default]{babel} 
\babelfont{rm}{TeXGyreTermesX} 
\babelprovide[import]{hindi}
\babelfont[*devanagari]{rm}{Lohit Devanagari}

\babelprovide[import]{chinese} 
\babelfont[*chinese]{rm}{Noto Serif CJK SC}

\babelprovide[import]{arabic}
\babelfont[*arabic]{rm}[
  UprightFont=NotoSansArabic.ttf
]{NotoSansArabic}

\babelprovide[import]{bengali}
\babelfont[bengali]{rm}[
  UprightFont=NotoSerifBengali.ttf
]{NotoSerifBengali}

\usepackage{microtype}
\usepackage{enumitem}
\usepackage{inconsolata}
\usepackage{tabularx}
\usepackage{graphicx}

\title{\textbf{\gradientRGB{XCR-Bench}{180,0,50}{0,0,180}}: Benchmarking Cross-Cultural Reasoning in \\LLMs via Culture-Specific Items and Hall's Triad}
\author{
\textbf{Mohsinul Kabir}$^{1,2}$,
\textbf{Tasnim Ahmed}$^{3}$,
\textbf{Md Mezbaur Rahman}$^{4}$,\\
\textbf{Shaoxiong Ji}$^{5,6}$,
\textbf{Hassan Alhuzali}$^{7}$,
\textbf{Yuechen Jiang}$^{1}$,\\
\textbf{Jimin Huang}$^{1,8}$,
\textbf{Sophia Ananiadou}$^{1,2}$\\
$^{1}$Department of Computer Science, National Centre for Text Mining,\\ The University of Manchester \quad
$^{2}$ELLIS Manchester\\
$^{3}$School of Computing, Queen's University, Ontario, Canada\\
$^{4}$Computer Science, University of Illinois Chicago\\
$^{5}$ELLIS Institute Finland \quad
$^{6}$University of Turku\\
$^{7}$Department of Computer Science and Artificial Intelligence,\\ Umm Al-Qura University, Makkah, Saudi Arabia,
$^{8}$The Fin AI\\
\texttt{ \textbf{Correspondence:} \{mdmohsinul.kabir, sophia.ananiadou\}@manchester.ac.uk}
}
\begin{document}
\maketitle

\begin{abstract}
Cross-cultural competence in large language models (LLMs) requires understanding and adapting Culture-Specific Items (CSIs) across varying cultural contexts. However, progress in evaluating this capability remains limited by the lack of high-quality CSI-annotated corpora with parallel cross-cultural sentence pairs. We introduce \textsc{XCR-Bench}, a Cross\textbf{(X)}-\textbf{C}ultural \textbf{R}easoning \textbf{Bench}mark containing \textbf{4.1k} parallel sentences and \textbf{1{,}098} CSIs across three reasoning tasks. \textsc{XCR-Bench} integrates Newmark's CSI framework with Hall's Triad of Culture, enabling evaluation across levels of cultural visibility- from observable practices to implicit social norms and values. Experiments on eight multilingual LLMs show that state-of-the-art models exhibit consistent weaknesses in identifying and adapting specific categories of CSIs, revealing a gap between surface-level recall and explicit cultural reasoning. Performance declines significantly on culturally sensitive categories and deeper cultural levels ($p<0.005$, 8/8 models), and adaptation quality varies systematically across target cultures and Bengali regional variants, indicating encoded regional and ethno-religious biases even within a single linguistic setting. We publicly release the corpus and code to support future research on cross-cultural NLP.\footnote{Code \& Corpus: \url{https://github.com/mohsinulkabir14/xcr_bench}}

\end{abstract}



\begin{figure}[ht]
  \includegraphics[width=\columnwidth]{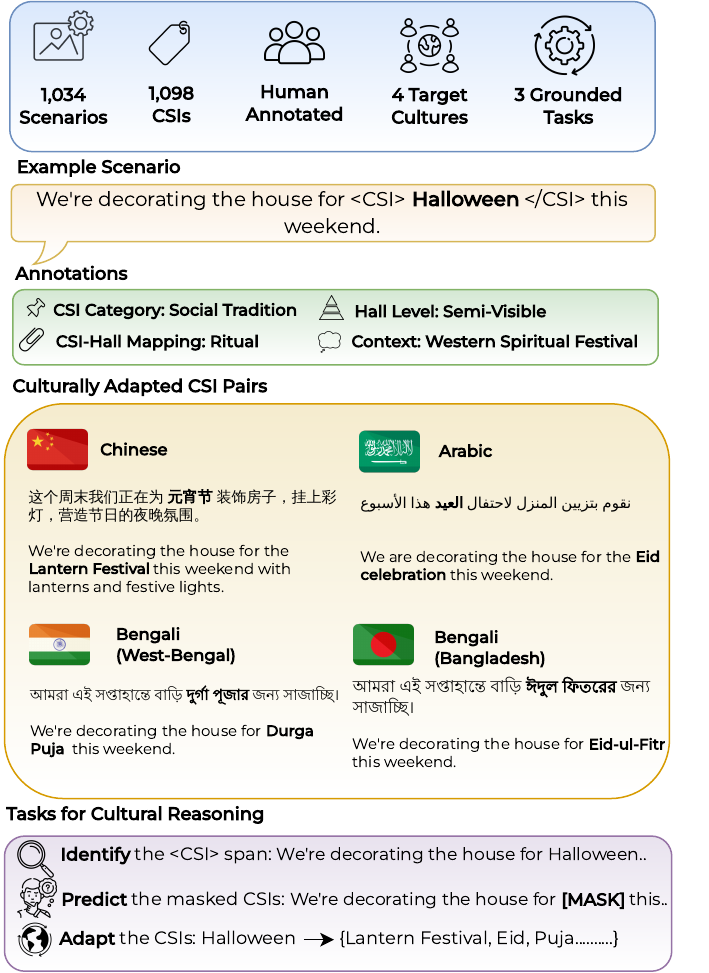}
  \caption{Overview of \textsc{XCR-Bench}: Western-source scenarios annotated with Newmark's CSI categories and Hall's cultural levels, adapted across four cultures (flag of Saudi Arabia reflects annotator background) in both intra- and inter-lingual settings, supporting three tasks: CSI Identification, Prediction, and Adaptation}
  \label{fig:xcr_intro}
\end{figure}

\section{Introduction}

\begin{table*}[ht]
\centering
\begin{adjustbox}{width=\textwidth}
\begin{tabular}{l l c c c c}
\toprule
\textbf{Dataset} &
\textbf{Domain} &
\textbf{Task} &
\makecell{\textbf{CSI}\\\textbf{Annot.}} &
\makecell{\textbf{Hall's}\\\textbf{Triad}} &
\makecell{\textbf{Cross-Cult.}\\\textbf{CSI Pair}} \\
\midrule

ApposCorpus~\citep{kementchedjhieva2020apposcorpus}
& Person, Org. & Trans. & \xmark & \xmark & \xmark \\

Adaption~\citep{peskov2021adapting}
& Celebrity & Trans. & \xmark & \xmark & \xmark \\

IdiomKB~\citep{li2024translate}
& Idioms & Trans. & \xmark & \xmark & \xmark \\

CulturalRecipes~\citep{cao2024cultural}
& Recipes & Trans. & \xmark & \xmark & \xmark \\

NormAd~\citep{rao2025normad}
& Soc. Etiq. & Reason. & \xmark & \xmark & \xmark \\

CAMT~\cite{yao2024benchmarking}
& Cult. Cat. & Trans. & \cmark & \xmark & \pmark \\

\midrule
\textsc{XCR-Bench} \textbf{(ours)}
& Cult. Cat. & Reason. & \cmark & \cmark & \cmark \\
\bottomrule
\end{tabular}
\end{adjustbox}

\caption{Comparison of \textsc{XCR-Bench} with prior cross-cultural datasets. Abbreviations: Cultural (\textit{Cult.}), Category (\textit{Cat.}), Annotation (\textit{Annot.}), Translation (\textit{Trans.}), Reasoning (\textit{Reason.}). Symbols denote the level of support: \cmark\ indicates full support, \xmark\ indicates the absence of support, and \pmark\ denotes partial or limited support.}
\label{tab:xcrbench_comparison}
\end{table*}

Large Language Models (LLMs) have been widely documented to exhibit a pronounced \textit{Anglo-American}, or \textit{Western} cultural bias, predominantly reflecting values and perspectives associated with the US and UK \citep{johnson2022ghost, dunn2024pre, atari2023humans}. This bias is often attributed to the West-centric composition of both their training data and the annotators involved in their development, making it a predictable artefact of their training process \citep{kabir2025n, navigli2023biases}. However, recent work has challenged a common implicit assumption in this line of research: the presence of cultural bias does not necessarily entail genuine cultural competence within the corresponding cultural context \citep{saha2025reading}. This observation leads us to ask: \textbf{to what extent do current LLMs perform genuine culturally grounded reasoning, rather than surface-level representation matching driven by Western-centric training distributions, when interpreting and adapting Culture-Specific Items (CSIs) across cultures and levels of cultural depth?} 

Beyond performance within predominantly Western cultural settings, a more challenging problem is evaluating whether LLMs can reason about and adapt culturally grounded meaning across different societies and contexts \citep{hershcovich2022challenges}. Existing work mainly focuses on translation or narrow substitution tasks involving named entities, idioms, recipes, or dialogues \citep{peskov2021adapting,cao2024cultural,li2024translate}, while recent benchmarks remain limited to partial CSI annotation or surface-level cultural categorisation \citep{yao2024benchmarking}. A key missing component is a model of cultural \emph{depth}, like Hall's Triad of Culture \citep{hall1976beyond} that distinguishes \textit{Visible} (technical), \textit{Semi-visible} (formal), and \textit{Invisible} (informal) levels, where visible behaviours are shaped by underlying norms, values, and beliefs. Newmark's CSIs and Hall's levels can capture \emph{orthogonal} dimensions: \textit{what} cultural element is present versus \textit{how deeply} its meaning is embedded, yet no existing typology or annotated resource situates CSIs within Hall's framework. As shown in Table~\ref{tab:xcrbench_comparison}, current datasets therefore lack the combination of fine-grained CSI annotations, theoretically grounded cultural depth modeling, and parallel cross-cultural CSI pairs. Consequently, existing evaluations \textbf{cannot assess whether LLMs move beyond surface-level representation matching to perform culturally grounded reasoning across different levels of cultural depth.}

To bridge this gap, we introduce the Cross\textbf{(X)}-\textbf{C}ultural \textbf{R}easoning \textbf{Bench}mark (\textsc{XCR-Bench}), a benchmark for evaluating cultural reasoning in LLMs beyond translation-level accuracy. Grounded in Newmark's and Hall's cultural theories, \textsc{XCR-Bench} consists of 4.1k parallel sentences and 1,098 unique CSIs across four target cultural settings: Chinese, Arabic, Bengali (West Bengal and Bangladesh). The corpus provides human annotations for CSI categories, CSI--Hall mappings, Hall's cultural depth levels, and culturally adapted CSI pairs from English into target cultures. Dataset construction follows a multi-stage human annotation and adjudication pipeline involving native speakers and expert validation, with substantial inter-annotator agreement for both CSI category and Hall-level annotations. \textsc{XCR-Bench} supports three tasks: \textit{CSI Identification}, \textit{CSI Prediction}, and \textit{CSI Adaptation}, together with task-specific evaluation metrics designed to measure culturally grounded reasoning rather than surface-level pattern matching.

We evaluate 8 state-of-the-art multilingual LLMs (4 closed-source, 4 open-weight). Across all three tasks, models handle surface-level cultural cues far more reliably than the implicit norms and values that require explicit reasoning. In summary, our contributions are:

\begin{itemize}
    \item We introduce \textsc{XCR-Bench}, a human-annotated benchmark linking Newmark's CSIs with Hall's Triad of Culture through CSI categories, cultural depth levels, and cross-cultural adaptation pairs.
    
    \item We formulate cultural competence as a culturally grounded reasoning problem through three tasks: \textit{CSI Identification}, \textit{CSI Prediction}, and \textit{CSI Adaptation}, together with task-specific evaluation metrics beyond translation-based settings.
    
    \item We provide a systematic evaluation of cross-cultural reasoning in multilingual LLMs across CSI categories, cultural depth levels, target cultures, and regional variants.
\end{itemize}

Beyond our core evaluation, \textsc{XCR-Bench} provides a valuable resource for downstream applications including cross-cultural machine translation, cross-lingual transfer, and the design of culturally grounded evaluation protocols for LLMs.



\begin{figure*}[ht]
  \includegraphics[width=\textwidth]{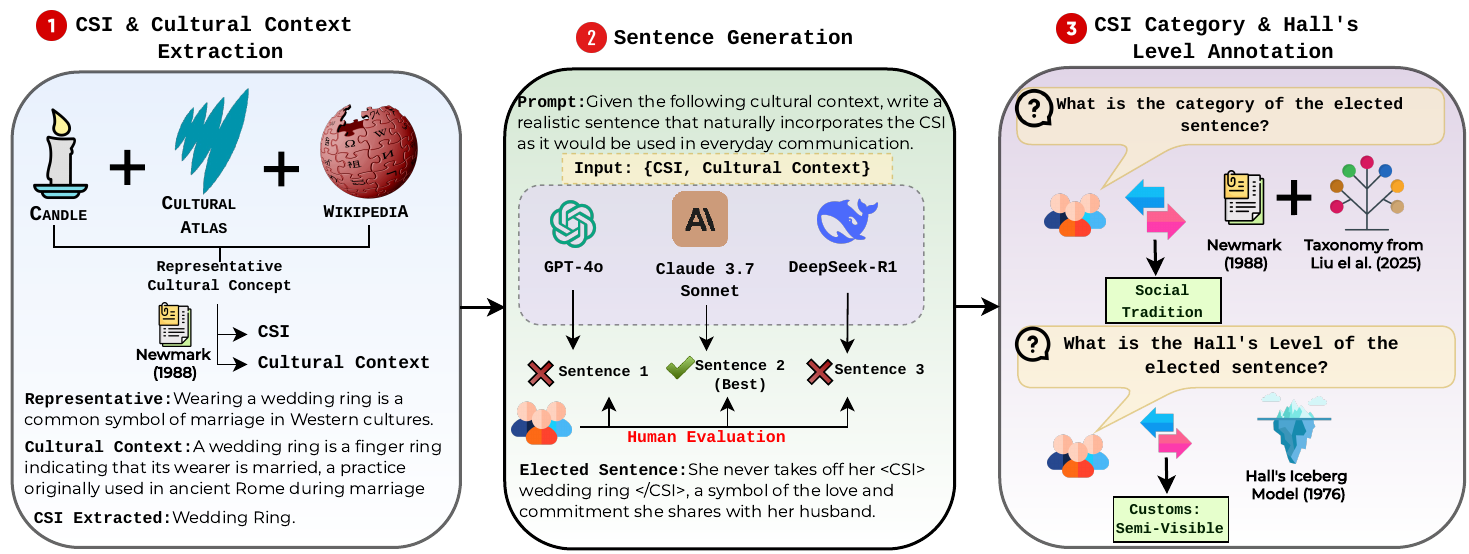}
  \caption{\textsc{XCR-Bench} corpus creation and annotation pipeline. (1) Culture-specific items (CSIs) and their contexts are extracted from well-known cultural databases. (2) Sentences are generated using LLMs and filtered for accuracy and fluency. (3) Each sentence is annotated with a CSI category and mapped to Hall's cultural levels.}
  \label{fig:xcr_bench}
\end{figure*}

\section{\gradientRGB{XCR-Bench}{180,0,50}{0,0,180} Corpus}
\label{sec:xcr_bench_corpus}
Due to the nuanced nature of CSIs, we incorporate human annotation and validation throughout \textsc{XCR-Bench} construction. The process comprises two phases: (1) building the base corpus with Western CSIs annotated by CSI categories and Hall's cultural levels; and (2) annotating adaptable CSI pairs for four target cultures (Chinese, Arabic, Bengali from West Bengal and Bangladesh) to form a parallel corpus. 

\subsection{Construction of Base XCR-Bench Corpus}

Figure~\ref{fig:xcr_bench} provides an overview of the base \textsc{XCR-Bench} construction pipeline. It includes: (1) CSI and cultural context extraction, (2) sentence generation with the extracted CSI, and (3) annotating the CSI to produce the category and Hall's level. 

\subsubsection*{CSI \& Cultural Context Extraction}

The first step in corpus construction involves identifying and extracting Culture-Specific Items (CSIs) representative of Western cultures. To this end, we leverage two structured cultural knowledge resources: {\Large C}{\normalsize ANDLE}\footnote{https://commonsense.scads.ai/candle/}
 and Cultural Atlas\footnote{https://culturalatlas.sbs.com.au/}. CANDLE provides cultural commonsense assertions, while Cultural Atlas offers curated descriptions of multicultural norms (details in Appendix~\ref{appendix:dataset_properties}).
We crawl cultural facts from these platforms, primarily focusing on the US and UK, supplemented by a selected few highly relevant examples from Europe. We then manually extract CSIs by mapping these statements to the CSI categories proposed by \citet{newmark1988textbook} (described in Appendix \ref{appendix:newmark_categories}). For each extracted CSI, we then retrieve a high-level cultural definition from Wikipedia to contextualize its usage. This process yields a set of CSIs paired with their corresponding cultural contexts.

\subsubsection*{Sentence Generation Incorporating CSIs}

In the second step, we generate realistic sentences that reflect daily communication by prompting GPT-4o, Claude-3.7-Sonnet, and DeepSeek-R1 with each CSI and its corresponding cultural context, as illustrated in Step~2 of Figure~\ref{fig:xcr_bench}. Each model produces a candidate sentence, resulting in three alternatives per CSI. Two human annotators then evaluate these candidates based on three criteria: \textit{accuracy}, \textit{fluency}, and \textit{realism}, and select the highest-scoring one. We explicitly emphasize \textit{realism} in our prompt and selection, as real-world cross-cultural contexts are inherently complex and essential for assessing cross-cultural competence \citep{rauba2024context}. This process yields a set of culturally grounded, naturalistic sentences aligned with the original CSI-context pairs. For experimental convenience, we annotate the selected sentence by enclosing the CSI within \texttt{<CSI>} and \texttt{</CSI>} tags.

\subsection*{CSI Category and Hall's Level Annotation}

Finally, we categorize each CSI based on its cultural context and the generated sentence from Steps~1 and~2, rather than at extraction time, as a CSI's category under Newmark's taxonomy can shift with contextual usage. For instance, \textit{wearing jeans} may fall under Material Culture or Social Culture depending on whether the context emphasises attire or workplace norms. Although we use Newmark's CSI categories as a foundational framework, their broad scope often merges diverse cultural elements. To address this, we incorporate the cultural taxonomy from \citet{liu2025culturally}, mapping its \textit{ideational} and \textit{social} dimensions to Newmark's Categories~3 (Social Culture) and~4 (Organizations, Customs, and Ideas), respectively. Annotations at this stage achieve a Cohen’s kappa of $0.68$, indicating substantial agreement between the two annotators.

We further annotate each CSI with its corresponding Hall cultural level, independently of Newmark's categorization: for instance, food-related CSIs may encode formal or informal cultural meanings without functioning as material artefacts. This context-sensitive treatment aligns with Skopos Theory \citep{schaffner1998}, which argues that the function of an element in the target communicative situation, rather than its surface form, determines its cultural role. Accordingly, we assign each CSI one of Hall's three levels of cultural visibility (e.g. Visible, Semi-visible, Invisible). The mapping of cultural elements to Hall's Triad is provided in Table \ref{tab:hall_visibility_mapping}. Annotation of Hall's cultural elements yields a Cohen's kappa of $0.64$ between the two annotators.

\subsection{Construction of the Parallel Corpus}

Each instance in our dataset is annotated with culturally adapted counterparts in four target cultures: Chinese, Arabic, and Bengali (two variants). The objective is to construct culturally equivalent instances of the Western CSIs from the base \textsc{XCR-Bench} corpus. Because cultural equivalence is inherently subjective and cannot be treated as a categorical labeling problem, we adopt an adjudicated annotation protocol \citep{klie2024analyzing}. 

Since multiple culturally valid adaptations may exist for a single CSI, we interpret variation as cultural pluralism rather than disagreement \citep{berry1974psychological}.
Consequently, conventional agreement-based metrics (e.g., IAA) are not appropriate for this setting.

Two native annotators per target culture independently produce adaptations following detailed guidelines (Appendix~\ref{appendix:annotation_guideline}), with disagreements resolved through discussion and final adjudication by an expert annotator with cultural and anthropological expertise (full procedure and qualifications in Section~\ref{appendix:annotation_procedure}). Following Newmark's framework for cultural equivalence \citep{newmark1988textbook}, annotators select one of four equivalence types: (1) a direct equivalent; (2) a functionally similar equivalent; (3) a neutral expression when no equivalent exists; or (4) \textit{Non-transferable} when the CSI conflicts with local norms (e.g., \texttt{<CSI>dating app</CSI>} in certain Arab contexts). Annotators are also provided with relevant cultural information from \textsc{Candle} and Cultural Atlas, and encouraged to modify sentence structure to ensure coherent, culturally appropriate intra- and inter-lingual adaptations.

\subsection{Dataset Analysis}
\label{sec:dataset_analysis}

The \textsc{XCR-Bench} corpus comprises $4{,}136$ parallel sentences spanning $4$ distinct cultures and $2$ regional variants. It contains $1{,}098$ unique Culture-Specific Items (CSIs), distributed across $7$ CSI categories. Table~\ref{tab:xcrbench_comparison} situates \textsc{XCR-Bench} in relation to existing datasets for culture-aware benchmarking. Most prior datasets in this area primarily provide translation pairs, with CAMT \citep{yao2024benchmarking} being the closest in scope, as it includes cross-cultural CSI pairs for a subset of its data. In contrast, \textsc{XCR-Bench} supports both reasoning-oriented and culturally aware adaptation evaluations by offering paired \textit{intra-lingual} and \textit{inter-lingual} annotations. Notably, \textsc{XCR-Bench} is the first dataset aligned with Hall's triad of culture, with explicit emphasis on the \textit{Semi-visible} and \textit{Invisible} layers, enabling evaluation of cultural reasoning beyond translation-focused competence.

\section{Cross-Cultural Reasoning Evaluation}


\subsection{CSI Identification}
\label{sec:csi_iden}
For the CSI Identification task, we remove the \texttt{<CSI>} tags from the base \textsc{XCR-Bench} sentences and prompt LLMs to identify the term(s) that are specific to Western (specifically US/UK) culture and salient for cross-cultural communication. To ensure a consistent interpretation of the task, we provide the definition of Culture-Specific Items (CSIs as defined by \citet{newmark1988textbook}) within the prompt during inference.

\subsection{CSI Prediction}
\label{sec:csi_pred}
For the CSI Prediction task, we replace the CSI spans in the base \textsc{XCR-Bench} sentences with a \texttt{[MASK]} token enclosed within \texttt{<CSI>} tags, and prompt models to predict culturally appropriate terms that reflect Western (US/UK) cultural usage. As in the \textit{CSI Identification} task, we include a brief definition of CSIs in the prompt to ensure a consistent interpretation of the task.

\begin{figure*}[ht]
    \centering
    \begin{subfigure}{0.49\textwidth}
        \centering
        \includegraphics[width=\textwidth]{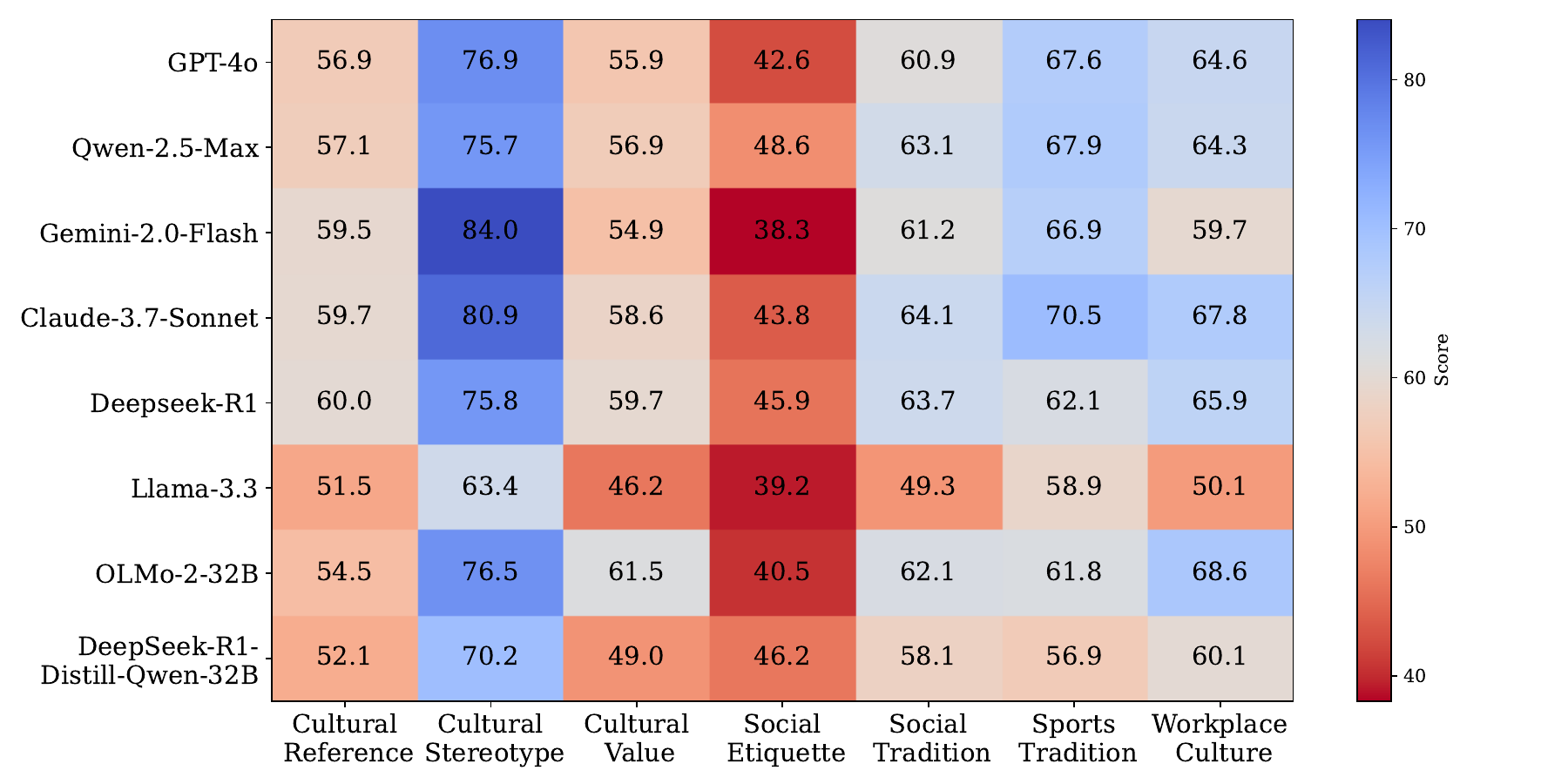}
        \caption{CSI Identification (SI-CSI)}
        \label{fig:csi_iden_heat}
    \end{subfigure}
    \begin{subfigure}{0.49\textwidth}
        \centering
        \includegraphics[width=\textwidth]{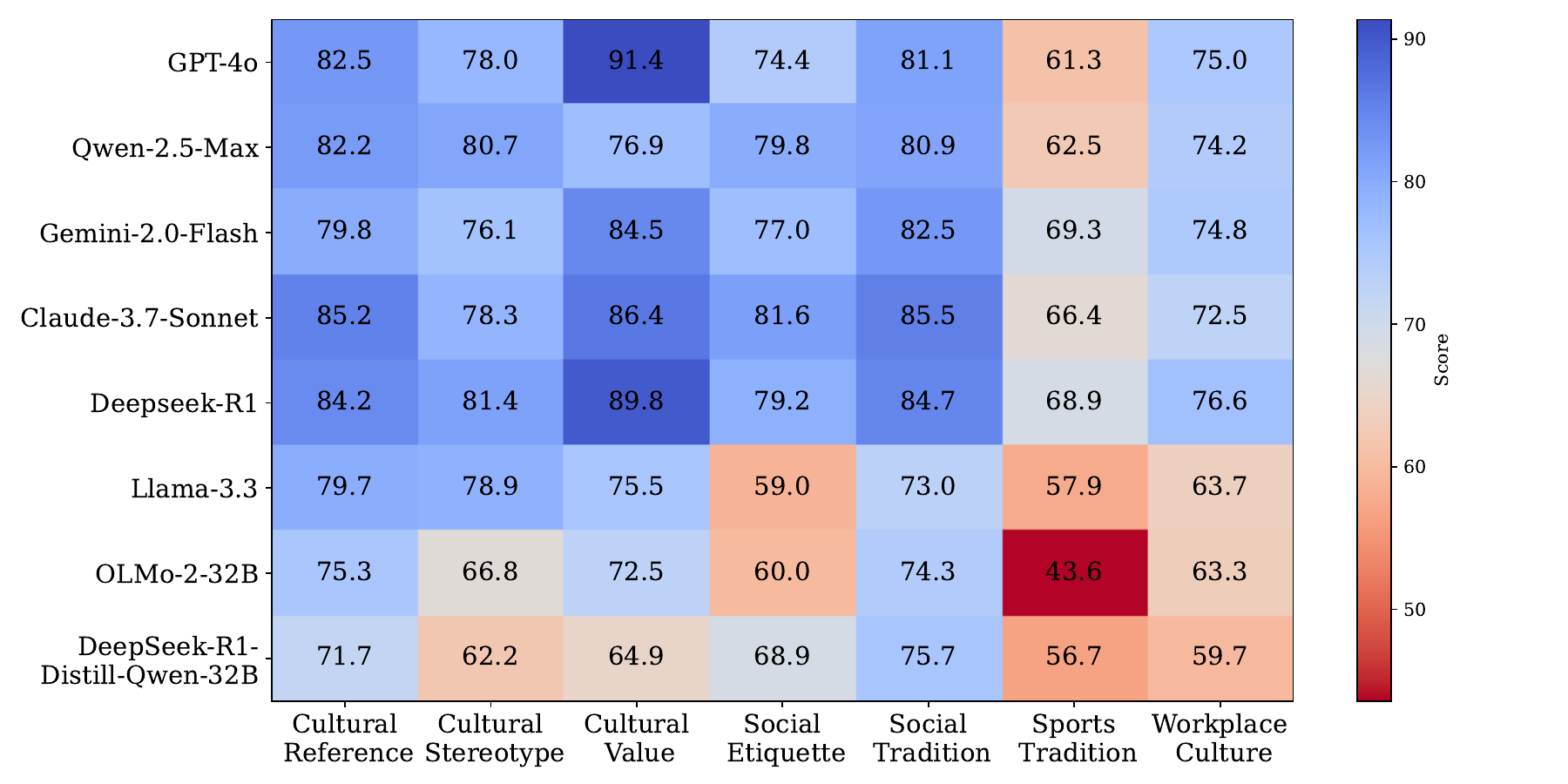}
        \caption{CSI Prediction (SP-CSI)}
        \label{fig:csi_pred_heat}
    \end{subfigure}

    \caption{Performance of LLMs across CSI categories for CSI Identification and Prediction tasks. For clarity, only soft evaluation metrics are visualized. The results reveal consistent difficulties in CSI Identification across most CSI categories.}
    \label{fig:csi_iden_adpt_heatmaps}
\end{figure*}

\subsection{CSI Adaptation}

In the CSI Adaptation task, we present LLMs with the base \textsc{XCR-Bench} sentences containing explicitly marked \texttt{<CSI>} spans and prompt them to adapt these CSIs to Arabic, Bengali, or Chinese cultural contexts. For each instance, models are required to produce both \textit{intra-lingual} (English) and \textit{inter-lingual} (target language) adaptations. In addition, we ask models to explicitly indicate which of Newmark's twelve cultural adaptation strategies (e.g., \textit{transference}, \textit{equivalence}, etc. full details in Appendix \ref{appendix:adaptation_procedure_newmark}) they employ when adapting each CSI.


Our prompts used for each task are presented in Appendix \ref{sec:prompts_appendix}.

\subsection{Evaluation Metrics}
We evaluate model performance on the CSI Identification and Prediction tasks using a combination of strict (``hard'') and lenient (``soft'') metrics designed to capture different aspects of correctness. The hard metrics penalize any deviation from the ground truth, while the soft metrics provide graded credit for semantically or superficially similar outputs, which is crucial for culturally nuanced tasks. In the Cultural Adaptation task, direct string matching is insufficient, as valid adaptations may use contextually congruent alternatives rather than a literal match. We therefore evaluate at two complementary levels using BERTScore: CSI$_{\text{bert}}$ assesses the semantic quality of individual adapted terms (within \texttt{<CSI>} tags), while SENT$_{\text{bert}}$ measures overall meaning preservation of the entire adapted sentence. A summary of all metrics is provided in Table~\ref{tab:eval-metrics} with detailed definitions and computation provided in Appendix \ref{appendix:detailed_evaluation_metrics}.

\begin{table}[ht]
\centering
\small
\begin{tabular}{@{}l l p{4.2cm}@{}}
\toprule
\textbf{Task} & \textbf{Metric} & \textbf{Description} \\
\midrule
\multirow{2}{*}{Identification} 
& \textbf{HI-CSI} & Exact string match between gold and predicted spans. \\
& \textbf{SI-CSI} & Normalised Levenshtein similarity with Hungarian alignment. \\
\midrule
\multirow{2}{*}{Prediction} 
& \textbf{HP-CSI} & Case-insensitive exact match. \\
& \textbf{SP-CSI} & Cosine similarity over Sentence-BERT embeddings. \\
\midrule
\multirow{2}{*}{Adaptation} 
& \textbf{CSI$_\text{bert}$} & BERTScore F1 over aligned CSI spans. \\
& \textbf{SENT$_\text{bert}$} & BERTScore F1 over full sentences. \\
\bottomrule
\end{tabular}
\caption{Evaluation metrics across the three tasks. Hard metrics (HI/HP) enforce strict correctness; soft metrics (SI/SP, CSI/SENT$_\text{bert}$) give graded credit. Formal definitions in Appendix~\ref{appendix:detailed_evaluation_metrics}.}
\label{tab:eval-metrics}
\end{table}

\subsection{Models and Prompting Technique} 

We evaluate a diverse set of recent multilingual LLMs, including both proprietary and open-source models: GPT-4o, Qwen-2.5-Max, Gemini-2.0-Flash, Claude-3.7-Sonnet, DeepSeek-R1 (671B), Llama-3.3 (70B), OLMo-2-32B, and DeepSeek-R1-Distill-Qwen-32B. All models are evaluated across the full suite of tasks. For prompting, we adopt part of the Expert Prompting strategy proposed by \citet{xu2023expertprompting}, which has been shown to elicit higher-quality responses than vanilla prompting.

\section{Results and Analysis}


\subsection{CSI Identification and Prediction}

 Table~\ref{tab:csi_west_results} reports the results of the Identification and Prediction tasks. While the two tasks employ different evaluation metrics and are not directly comparable on absolute scores, the qualitative profile of model behaviour differs significantly. Prediction resembles a mask-filling task, which is a setting that is well-aligned with autoregressive pretraining over text containing Western CSIs. In contrast, CSI Identification requires explicit reasoning about which elements in a sentence are culturally specific and salient for cross-cultural communication, a substantially more demanding capability that all models struggle with. Qualitative inspection of model outputs across GPT-4o, Llama-3.3, Qwen-2.5-Max, and Gemini-2.0-Flash reveals a consistent failure mode: models tend to mark surface-level entities as CSIs while overlooking the culturally salient behaviour they participate in. For instance, given ``He is the CEO, but everyone calls him by his \texttt{<CSI>first name</CSI>} John'', models correctly fill the masked span with \textit{first name} in Prediction, yet mark \textit{CEO} as the CSI in Identification, failing to recognise that addressing superiors by first name (rather than the title itself) is the culturally salient behaviour, common in Western workplaces but taboo in many others. The recurrence of this pattern across model families supports our claim that CSI Identification requires explicit cultural reasoning and judgment beyond surface lexical recall.

\begin{table}[ht]
\centering
\resizebox{\columnwidth}{!}{
\begin{tabular}{lcccc}
\toprule
\textbf{Model} & \multicolumn{2}{c}{\textbf{CSI Identification}} & \multicolumn{2}{c}{\textbf{CSI Prediction}} \\
 & HI-CSI & SI-CSI & HP-CSI & SP-CSI \\
\midrule
GPT-4o & 50.94 & 57.74 & 43.70 & 78.47 \\
Qwen-2.5-Max & \underline{54.99} & 59.75 & 43.70 & 79.30 \\
Gemini-2.0-Flash & 48.73 & 57.15 & 42.94 & 78.96 \\
Claude-3.7-Sonnet & 47.95 & \textbf{60.47} & \underline{46.76} & \textbf{81.76} \\
DeepSeek-R1 & \textbf{59.05} & \underline{60.01} & \textbf{47.33} & \underline{81.68} \\
Llama-3.3 & 52.62 & 48.76 & 34.54 & 69.90 \\
OLMo-2-32B & 44.91 & 57.71 & 31.87 & 68.30 \\
DeepSeek-R1-Distill-Qwen-32B & 51.34 & 55.14 & 29.96 & 69.49 \\
\bottomrule
\end{tabular}
}
\caption{CSI Identification and Prediction performance across models. Highest and second-highest scores are bolded and underlined, respectively. Values are percentages; higher is better.}
\label{tab:csi_west_results}
\end{table}

\textbf{Model performance is highly category-dependent, with systematic strengths and weaknesses that diverge across the two tasks.} Figure~\ref{fig:csi_iden_adpt_heatmaps} illustrates category-wise trends. In CSI Identification (Figure~\ref{fig:csi_iden_heat}), models perform best on \textit{Cultural Stereotype} (mean $\approx 75$) and worst on \textit{Social Etiquette} (mean $\approx 43$), with \textit{Cultural Reference} and \textit{Cultural Value} also posing consistent difficulty. In CSI Prediction (Figure~\ref{fig:csi_pred_heat}), the profile shifts: \textit{Cultural Value} and \textit{Cultural Reference} become the strongest categories, while \textit{Sports Tradition} drops to the lowest. This divergence indicates that the categories LLMs find easiest to \emph{predict} from context are not the same as those they reliably \emph{identify} as culturally specific, suggesting that strong pretraining exposure to a category enables mask-fill recovery without conferring the explicit cultural-salience reasoning required for identification. 

\textbf{Statistical analysis.} A Friedman test confirms significant performance differences across CSI categories for both Identification ($\chi^2(6) = 43.29$, $p < 0.001$) and Prediction ($\chi^2(6) = 37.55$, $p < 0.001$). Post-hoc Nemenyi tests (Table~\ref{tab:nemenyi_appendix}) reveal markedly different difficulty profiles: in Identification, \textit{Cultural Stereotype} is the easiest and \textit{Social Etiquette} the hardest, whereas in Prediction, \textit{Sports Tradition} and \textit{Workplace Culture} become the hardest while \textit{Cultural Reference}, one of the hardest in Identification, becomes among the easiest. This reversal indicates that the two tasks tap qualitatively different capabilities.

\begin{figure}[ht]
  \includegraphics[width=\columnwidth]{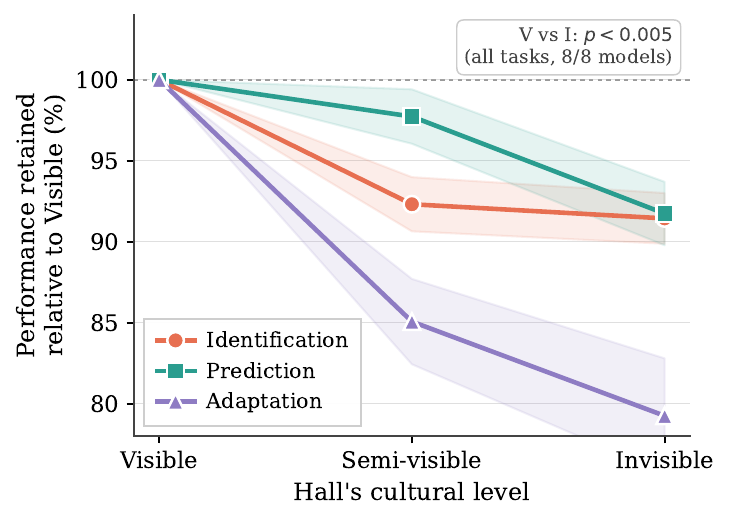}
\caption{Within-task performance retention across Hall's cultural triad. For each model and task, Semi-visible and Invisible scores are normalised to its own Visible-level score (=100\%) using the task's native metric (SI-CSI for Identification, SP-CSI for Prediction, CSI$_\mathrm{bert}$ for Adaptation); lines show mean across 8 LLMs with $\pm 1$ SE. The figure compares relative degradation patterns, not absolute scores. Visible-vs-Invisible decline is significant in all tasks (one-sided Wilcoxon on raw scores, $p < 0.005$, 8/8 models).}
  \label{fig:hall_decline}
\end{figure}

\textbf{Performance systematically declines from Visible to Invisible levels of Hall's cultural triad in both tasks.} Figure~\ref{fig:hall_decline} shows Visible-vs-Invisible decline for both tasks (full result in Table~\ref{tab:hall_identification_prediction}), and Table~\ref{tab:hall_category_csi_iden} reports per-category performance across Hall's levels. Models perform substantially worse on semi-visible and invisible elements such as \textit{Appropriacy} and \textit{Orientations}, which encode norms, values, and beliefs rather than observable practices. Together with the category-wise findings, these results indicate that LLMs handle surface-level cultural cues considerably more reliably than the deeper, norm- and value-driven aspects of culture that support genuine cross-cultural competence.

\begin{table}[ht]
\centering
\resizebox{\columnwidth}{!}{
\begin{tabular}{l*4{c}:*4{c}}
\toprule
& \multicolumn{2}{c}{\textbf{Arabic}}
& \multicolumn{2}{c}{\textbf{Chinese}}
& \multicolumn{2}{c}{\textbf{Bengali (WB)}}
& \multicolumn{2}{c}{\textbf{Bengali (BD)}} \\
\cmidrule(lr){2-3}\cmidrule(lr){4-5}\cmidrule(lr){6-7}\cmidrule(lr){8-9}
Model & Intra & Inter & Intra & Inter & Intra & Inter & Intra & Inter \\
\midrule
DeepSeek-R1
& 28.75 & 65.86
& \underline{33.93} & 71.34
& \textbf{37.23} & \textbf{48.44}
& \textbf{35.40} & \textbf{47.41} \\
Gemini-2.0-Flash
& 27.78 & \underline{67.09}
& 32.74 & 70.33
& 26.44 & 44.20
& 25.94 & 43.13 \\
GPT-4o
& \textbf{35.21} & 67.94
& \textbf{38.85} & \underline{72.42}
& 35.46 & 45.06
& 32.42 & 44.14 \\
Llama-3.3
& 28.93 & 66.25
& 34.60 & 72.10
& \underline{37.07} & \underline{48.27}
& \underline{35.17} & \underline{47.19} \\
Qwen-2.5-Max
& 33.21 & 67.66
& 34.67 & \textbf{72.74}
& 25.42 & 46.59
& 23.26 & 44.54 \\
OLMo-2-32B
& 28.96 & 66.26
& 34.85 & 72.03
& 23.61 & 43.32
& 21.93 & 42.02 \\
Claude-3.7-Sonnet
& \underline{33.47} & \textbf{68.69}
& 36.88 & 72.07
& 34.10 & 46.01
& 30.67 & 42.86 \\
D-R1-D-Qwen-32B
& 23.96 & 59.49
& 34.48 & 69.73
& 28.25 & 36.54
& 26.13 & 35.46 \\
\midrule
Average
& 30.03 & 66.16
& 35.12 & 71.59
& 30.95 & 44.80
& 28.87 & 43.34 \\
\bottomrule
\end{tabular}
}
\caption{CSI adaptation performance (\%) across models measured using the CSI$_\mathrm{bert}$ metric, reported in both intra-lingual and inter-lingual settings across four target cultures. Sentence-level scores (SENT$_\mathrm{bert}$) are reported in Table \ref{tab:sentbert_adaptation_result} in Appendix~\ref{appendix:additional_results}. Abbreviations: Bangladesh (\textit{BD}), West Bengal (\textit{WB}).}
\label{tab:csi_adaptation_result}
\end{table}

\subsection{Cultural Adaptation}
The CSI Adaptation task evaluates LLMs' ability to adapt culture-specific content in both intra-lingual and inter-lingual settings (Table~\ref{tab:csi_adaptation_result}). We focus on CSI$_\mathrm{bert}$, as adaptation typically involves localised substitutions that leave overall sentence structure largely unchanged, making SENT$_\mathrm{bert}$ less discriminative. Across cultures, LLMs achieve highest adaptation performance for Chinese, followed by Arabic, with Bengali consistently lowest across models and CSI categories.

\begin{figure}[ht]
  \includegraphics[width=\columnwidth]{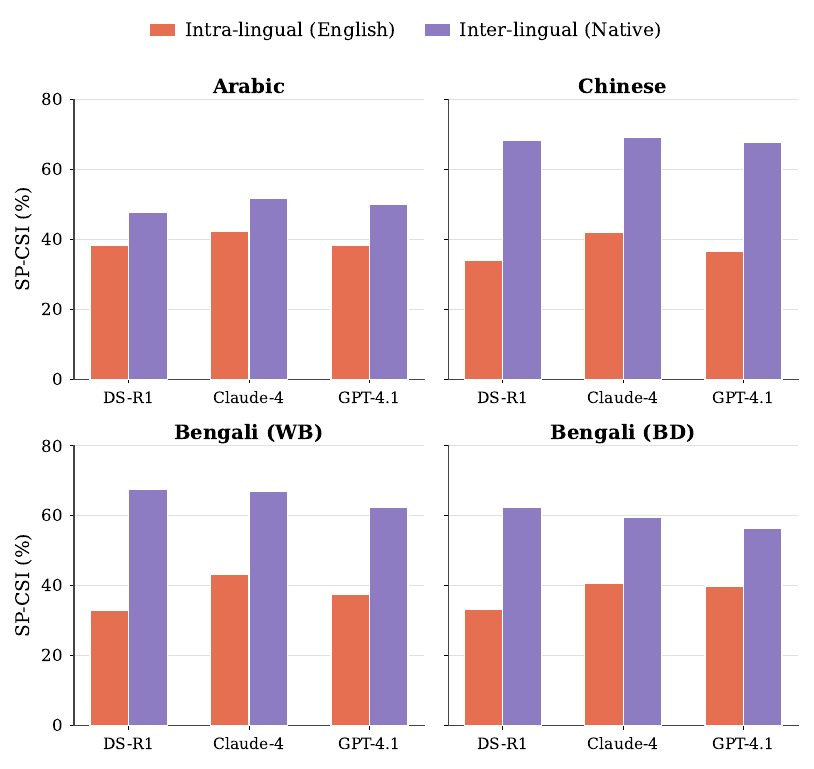}
\caption{CSI Prediction performance (SP-CSI) under \textit{Intra-lingual (English)}: English source; and \textit{Inter-lingual (Native)}: native-language source. Models: DS-R1 (DeepSeek-R1), Claude-4 (Claude-4-Sonnet), GPT-4.1.}
  \label{fig:intra_vs_inter}
\end{figure}

\textbf{LLMs consistently exhibit stronger CSI adaptation performance in inter-lingual settings than in intra-lingual settings.}
Across all four cultures, inter-lingual adaptation outperforms intra-lingual in 8/8 models (one-sided Wilcoxon signed-rank test, $p < 0.005$ for each culture), with mean CSI$_\mathrm{bert}$ gains of $+36.1$ (Arabic), $+36.5$ (Chinese), $+13.9$ (Bengali-WB), and $+14.5$ (Bengali-BD) percentage points. This pattern suggests that while LLMs are relatively effective at adapting Western cultural content when operating directly to the target language, they encounter greater difficulty when adapting non-Western cultural concepts expressed in English. To better understand where these gains arise, we analyse performance differences at the CSI category level. Figures~\ref{fig:inter_vs_intra_arabic_chinese} and~\ref{fig:inter_vs_intra_wb_bd} show that inter-lingual adaptation yields the largest improvements in the \textit{Cultural Stereotype} category and the smallest gains in \textit{Cultural Reference}, suggesting that LLMs more readily adapt stereotypical or conventionalised cultural knowledge in the target language, while culturally grounded references tied to specific contexts or entities remain more challenging. A similar Visible-to-Invisible decline emerges through Hall's triad, though less pronounced in Bengali  (Table~\ref{tab:hall_adaptation}).

\textbf{Disentangling language and cultural context.} To test whether inter-lingual gains reflect cultural reasoning rather than target-language fluency, we conduct a controlled ablation on a 5\% sample for CSI Prediction across all four target cultures, evaluating DeepSeek-R1, GPT-4.1, and Claude-4-Sonnet.\footnote{GPT-4o and Claude-3.7-Sonnet, used in our main experiments, were deprecated during the ablation period.} As shown in Figure~\ref{fig:intra_vs_inter}, native-language source inputs yield substantially stronger culturally grounded predictions in Chinese and both Bengali variants ($+20$ to $+30$ points on SP-CSI), with smaller gains for Arabic. We attribute this gap to a representational mismatch in intra-lingual prediction: when asked to insert culturally appropriate items into an English sentence, models often default to surface-level cultural markers such as transliterations or stereotypical borrowings (e.g., \textit{ni hao}, \textit{siheyuan}). Inter-lingual prediction, in contrast, lets models operate directly within the target language's cultural-pragmatic space, where culturally grounded expressions are more strongly represented in training data. A notable secondary observation arises in Bengali, which has no gendered third-person pronoun (only \foreignlanguage{bengali}{`সে'} for he/she): when the English source contains \textit{he} or \textit{she}, models often introduce a male-default term in the predicted CSI, but this gendered cue is lost in the inter-lingual setting where the source is already in Bengali.

\textbf{Adaptation Procedure.}
During CSI adaptation, we prompt LLMs to indicate which of twelve adaptation strategies (Appendix~\ref{appendix:adaptation_procedure_newmark}) is employed. Models frequently select \textit{Couplet} combinations, with \textit{Cultural Equivalent} the most common single strategy. Per-method effectiveness is reported in Table~\ref{tab:best_worst_adaptation_methods}.

\textbf{Regional Bias in Adaptation.}
Bengali constitutes a particularly informative case for evaluating regional and religious bias. Speakers are predominantly Muslim ($\sim$71\%) with a Hindu minority ($\sim$28\%), split between Bangladesh ($\sim$59\% of speakers) and West Bengal, India ($\sim$38\%) \citep{bsb2022}, and the two populations differ in religion, dialect (\textit{Bangal} vs. \textit{Ghoti}), and culturally salient terminology.


\begin{table}[ht]
\centering
\small
\resizebox{\columnwidth}{!}{
\begin{tabular}{lcc}
\toprule
\textbf{Model} &
\begin{tabular}[c]{@{}c@{}}\textbf{Intra-lingual} \\ \textbf{(SENT$_\text{bert}$ / CSI$_\text{bert}$)}\end{tabular} & 
\begin{tabular}[c]{@{}c@{}}\textbf{Inter-lingual} \\ \textbf{(SENT$_\text{bert}$ / CSI$_\text{bert}$)}\end{tabular} \\
\midrule
DeepSeek-R1
& $0.003 / 0.005$
& $-0.0002 / -0.003$ \\

Gemini-2.0-Flash
& $0.002 / 0.003$
& $0.001 / 0.012^{**}$ \\

GPT-4o
& $0.005 / 0.023^{*}$
& $-0.0002 / -0.001$ \\

Llama-3.3
& $0.003 / 0.006$
& $-0.0002 / -0.004$ \\

OLMo-2-32B
& $0.003 / 0.005$
& $-0.0002 / 0.009$ \\

Qwen-2.5-Max
& $0.001 / 0.012$
& $0.002 / 0.030^{**}$ \\

Claude-3.7-Sonnet
& $0.009 / 0.020^{*}$
& $0.004 / 0.026^{*}$ \\

DeepSeek-R1-Distill-Qwen-32B
& $0.005 / 0.005$
& $0.000 / 0.016$ \\
\bottomrule
\end{tabular}
}
\caption{Mean regional performance differences in Bengali cultural adaptation (West Bengal $-$ Bangladesh), aggregated across CSI categories. Statistical significance is evaluated using a one-sided Wilcoxon signed-rank test ($^{*}p<0.05$, $^{**}p<0.01$).}
\label{tab:hindu_muslim_mean_gap}
\end{table}

To assess whether such variation influences LLM-based cultural adaptation, we evaluate Bengali CSIs against both West Bengal and Bangladesh cultural corpora and compare adaptation performance across settings. As reported in Table~\ref{tab:hindu_muslim_mean_gap}, all statistically significant mean differences favour West Bengal over Bangladesh, indicating stronger model alignment with West Bengal–specific cultural knowledge, particularly for Claude and Gemini. While the effect sizes are not large, this reflects an inherent limitation of embedding-based metrics in low-resource settings: native speakers note that models systematically prefer lexical variants such as \foreignlanguage{bengali}{জল} (Hindu-majority usage) over \foreignlanguage{bengali}{পানি} (Muslim-majority usage), both denoting ``water''. Because these variants are semantically equivalent, Bangla BERT embeddings produce highly similar vectors, masking culturally meaningful preferences and compressing the observed effect size.

Native-speaker analysis confirms a consistent \textit{Ghoti}-leaning pattern of LLMs across multiple domains: everyday vocabulary (\foreignlanguage{bengali}{ভোজ} over \foreignlanguage{bengali}{খানা}, both denoting `feast'), kinship terms (\foreignlanguage{bengali}{দাদা} over \foreignlanguage{bengali}{ভাই}, both denoting `brother'), festival names (\foreignlanguage{bengali}{পুজো} over \foreignlanguage{bengali}{ঈদ} when adapting Christmas). These choices systematically favour \textit{Ghoti} lexical and cultural markers over \textit{Bangal} alternatives, despite Bangladesh-Bengalis constituting the demographic majority within the broader Bengali-speaking population. And more surprisingly, this arises in a low-resource setting where overall Bengali performance is comparatively weaker than other languages (Table~\ref{tab:csi_adaptation_result}). Together with the quantitative results, these findings suggest that LLMs can encode subtle regional and ethno-religious biases within a single language, favouring culturally dominant or more frequently represented variants in their training data.

\section{Conclusion}
We introduce \textsc{XCR-Bench}, a human-annotated parallel corpus with three distinct reasoning tasks, designed to support the evaluation of cross-cultural reasoning in large language models (LLMs). By integrating Newmark's concept of CSIs with Hall's triad of culture, \textsc{XCR-Bench} enables fine-grained analysis for subtle cultural elements. These annotations further enable mechanistic analysis of how cultural information is encoded within model internals. Our evaluation shows that contemporary LLMs struggle to identify and adapt CSIs in cross-cultural contexts. We also observe that cultural adaptation responses often encode regional and ethno-religious biases within a single linguistic setting. Our work provides high-quality data and empirical insights to advance the study of cross-cultural reasoning in NLP.

\section*{Limitations}
While our study introduces a novel benchmark for evaluating LLMs' cross-cultural adaptability, it is subject to several limitations.

\textbf{Limited Coverage of Newmark Categories.} Newmark's framework \citep{newmark1988textbook} outlines five categories of CSIs, as detailed in Appendix~\ref{appendix:newmark_categories}. As shown in Table~\ref{tab:csi_category_mapping}, the CSIs in \textsc{XCR-Bench} primarily span categories 2 and 3. This focus stems from our emphasis on the semi-visible and invisible levels of Hall's Iceberg model of culture, which encompass social values, beliefs, and norms. Consequently, our benchmark does not cover categories 1, 4, and 5 (e.g., ecology, material culture, and gestures), representing a gap that future work could address by expanding the CSI typology.

\textbf{Limited Coverage of Cultures.} Our corpus includes CSI annotations for only four distinct cultures (Western US/UK, Arabic, Bengali, and Chinese). While this selection enables targeted insights, incorporating additional languages and cultures would likely reveal broader patterns in LLMs' adaptation capabilities. This constraint was influenced by our available resources for collaboration and annotation.

\textbf{Limited Coverage of Prompting Techniques.} In our main experiments, we employed the expert prompting strategy from \citet{xu2023expertprompting}. To assess whether our findings are sensitive to prompt formulation, we conducted a preliminary ablation on the CSI Prediction task with DeepSeek-R1, Claude-4-Sonnet, and GPT-4.1 under three prompt variants: expert prompting, chain-of-thought reasoning, and few-shot prompting with three examples (Appendix~\ref{appendix:additional_results}, Figure~\ref{fig:prompt_sensitivity}). No prompting strategy consistently outperforms the others across models: few-shot prompting helps Claude-4-Sonnet, chain-of-thought marginally helps DeepSeek-R1 on the hard metric, while expert prompting remains strongest for GPT-4.1, and the qualitative patterns reported in our main results (e.g., the Visible-to-Invisible decline and category-wise weaknesses) hold across all three prompt formulations. This suggests that the cross-cultural reasoning gaps we observe are unlikely to be improved by prompt engineering alone, and instead reflect deeper limitations in how culturally grounded knowledge is represented during pretraining. A systematic study across all tasks, models, and a broader set of prompting strategies remains an important direction for future work, particularly because it may reveal how much of the cross-cultural reasoning gap can be closed through prompt-level interventions versus what requires model-intrinsic improvements such as targeted cultural pretraining or alignment.

\textbf{Annotator-Bound Cultural Representation.} While our annotators are native speakers with strong cultural familiarity, the adapted sentences inevitably reflect their specific cultural backgrounds and may not capture the full diversity within each target culture. This is particularly relevant for Arabic, which spans more than twenty countries with substantial inter-country variation in dialect, religious practice, and social custom, as well as intra-country diversity within individual nations \citep{naous2024having}. Our two Arabic annotators are based in Saudi Arabia, and their adaptations may therefore better reflect Gulf or Saudi-specific cultural norms than, for instance, North African or Levantine ones. Similar considerations apply to our Chinese and Bengali annotations, though we partially address regional variation in Bengali through separate West Bengal and Bangladeshi annotations. Broader coverage of intra-cultural diversity-through larger and more geographically distributed annotator pools—remains an important direction for future work.

\textbf{Western-Centric Source Framing.} \textsc{XCR-Bench} is constructed around Western (US/UK) source CSIs adapted into non-Western contexts, and does not currently evaluate non-Western-to-Western or non-Western-to-non-Western adaptation. This design is deliberate: it tests LLMs' Western cultural knowledge as a baseline while simultaneously evaluating their ability to recontextualise that knowledge into divergent cultural settings—a more challenging and practically consequential capability than knowledge recall alone, motivated by recent evidence that Western training-data bias does not guarantee strong Western cultural competence \citep{saha2025reading}. Our source resources (\textsc{Candle} and Cultural Atlas) further support this choice, offering well-structured Western cultural references with explicit cross-cultural mappings suitable for controlled evaluation. Nonetheless, this framing restricts cultural coverage, and findings may not fully generalise to settings where source items originate outside Western contexts. Extending the benchmark to non-Western-to-non-Western and non-Western-to-Western transfer is an important direction for future work.


\bibliography{custom}

\appendix
\section{Background Study}
\label{appendix:background_study}

\subsection{Measuring `Culture' in LLMs}

Culture is a multifaceted concept encompassing cultural heritage (art, music, food), interpersonal interaction norms, and the collective ways of life that distinguish groups from one another \citep{adilazuarda2024towards}. Sociologically, it has been described as the abstract patterns of ideas and principles that guide human behavior in practically effective ways. Recent studies have predominantly relied on global surveys to assess the cultural knowledge of large language models (LLMs) regarding specific countries. In particular, multiple-choice questions drawn from surveys such as the World Values Survey (WVS) \citep{haerpfer2020world}, Hofstede's Cultural Dimensions \citep{hofstede2010cultures}, or PEW Global Surveys \footnote{https://www.pewresearch.org/} are presented to LLMs to evaluate their understanding of particular cultures or countries \citep{zhao2024worldvaluesbench, durmus2023towards, kharchenko2024well}. \citet{wang2024cdeval} uses Hofstede's Survey to construct a benchmark by forming natural scenario-based questions from the questionnaire itself using GPT-4. While these surveys provide an excellent means of gathering real-world human perspectives, their use for evaluating cultural knowledge in LLMs has several limitations: (i) LLMs are restricted to the predefined choices in multiple-choice formats, yet culture is a nuanced concept that often requires more sophisticated and open-ended responses \citep{beugelsdijk2018dimensions}; (ii) altering the order of multiple-choice options can lead LLMs to produce entirely different answers, which undermines reliability for a multifaceted notion like culture \citep{pezeshkpour2024large}. In response, several studies have critiqued this approach and advocated for more open-ended evaluations to better capture the sophistication and intricacy of cultural understanding \citep{kabir-etal-2025-break, rottger2024political}. Other works have proposed specialized frameworks targeting specific cultural facets or proxies, such as food, etiquette, or regional differences, rather than attempting a holistic evaluation of culture \citep{naous2024having, adilazuarda2024towards}.

\subsection{Evaluation of Cross-Cultural Competence of LLMs}

The ability of large language models (LLMs) to adapt meanings across cultural contexts is essential for broadening access to AI systems and mitigating harms from culturally maladapted outputs \citep{kabir2025semantic}. To evaluate LLMs' proficiency in cross-cultural communication, recent studies have drawn on the concept of Culture-Specific Items (CSIs), as defined by Newmark \citep{newmark1988textbook}, emphasizing that identifying and adapting these items is critical for effective cross-cultural exchange.\\
Several works have explored CSI adaptation in specific domains. For example, \citet{singh2024translating} adapted American CSIs to Indian equivalents using the Friends Dialogues corpus \footnote{https://convokit.cornell.edu/documentation/friends.html}. Similarly, \citet{yao2024benchmarking} introduced CAMT, a parallel corpus across six languages featuring paired CSIs, to assess LLMs' cultural adaptability. In a domain-specific approach, \citet{cao2024cultural} proposed tasks for translating and culturally adapting recipes between Chinese- and English-speaking cuisines. However, these studies primarily rely on machine translation evaluations and often lack explicit CSI annotations, limiting their ability to probe models' targeted capabilities in CSI span identification and sentence-level adaptation.
More recent efforts have expanded the scope of cross-cultural evaluations. \citet{guo2025large} proposed a methodology using Cross-Cultural Core Concept Sets (CCCs) to assess LLMs' cross-cultural understanding in real-world scenarios, enabling more authentic capability assessments. \citet{wu2025socialcc} introduced SocialCC, a benchmark for evaluating cultural competence through 3,060 multi-turn interactive intercultural scenarios spanning 60 countries. Incorporating multimodality, \citet{song2025culture} developed C$^3$B, a benchmark with 2,000 images and over 18,000 QA pairs across three tasks of varying difficulty: basic visual recognition, higher-level cultural conflict understanding, and cultural content generation. In a related vein, \citet{kim2025tom} created a benchmark to evaluate biases in multimodal LLMs across high- and low-resource cultural settings.\\\
Our work complements these efforts by providing a culturally rich corpus of realistic scenarios, with explicit annotations for CSI categories, Hall's cultural levels, and cross-cultural CSI pairs, thereby advancing research in cross-cultural NLP.

\section{Utilizing Newmark's Theory of CSIs}

\subsection{Newmark's Definition of Culture and CSI Categories}
\label{appendix:newmark_categories}

Newmark \citep{newmark1988textbook} defines culture as a shared way of life, encompassing the distinctive characteristics, beliefs, and practices of a particular social group. As outlined in the cultural proxies in Appendix \ref{appendix:background_study}, these differences span multiple dimensions- from tangible elements like food and materials to abstract ones such as emotions and values \citep{thompson2020cultural, adilazuarda2024towards}. Newmark \citep{newmark1988textbook}, in the 9th Chapter of his book \textit{A textbook of Translation},  terms such concepts Culture-Specific Items (CSIs), whose identification and handling are vital for effective cross-cultural communication. He proposes five key categories of CSIs that are especially critical in cross-cultural translation: \\\\
\textbf{Category 1: Ecology} covers the relationship between a community and its natural environment, including indigenous flora, fauna, geography, and climate.\
\textit{Example:} Translating the specific name for a seasonal desert wind, such as the \textit{Simoom} in Arabic cultures, or the \textit{Khamsin} in Egypt.\\\\
\textbf{Category 2: Material Culture (Artefacts)} includes all tangible, human-made objects that shape or reflect a society's way of living, such as tools, clothing, architecture, and food.\
\textit{Example:} Translating terms like the \textit{Shamagh} (the traditional checked men's headdress in the Gulf region) or \textit{Iftar} (the meal to break the Ramadan fast).\\\\
\textbf{Category 3: Social Culture} refers to the patterns of social activity, including work, leisure, and communal practices that define daily and ceremonial life.\
\textit{Example:} Translating the name of a traditional social gathering like the \textit{Majlis} (a council or sitting room for community discussion in Arab cultures) or \textit{Jiāzǐ} (a 60th birthday celebration in Chinese culture).\\\\
\textbf{Category 4: Organizations, Customs, and Ideas} encompasses the abstract and institutional aspects of culture, including its political, legal, religious, artistic, and customary systems.\
\textit{Example:} Translating concepts such as \textit{Apritismo} (a Peruvian political ideology) or \textit{Barangay} (the smallest administrative division in the Philippines, rooted in pre-colonial history).\\\\
\textbf{Category 5: Gestures and Habits} involves non-verbal communication, body language, and routine behaviors whose meanings are culturally encoded and may not have direct equivalents.\
\textit{Example:} Interpreting gestures like the \textit{chin flick} (a sign of dismissal or negation in parts of Italy and France) or the \textit{"Moutza"} (an offensive hand gesture in Greece).

\subsection{CSI Categories Mapping}
\label{appendix:newmark_mapping}

\begin{table*}[ht]
\centering
\small
\setlength{\tabcolsep}{5pt}
\renewcommand{\arraystretch}{1.15}
\begin{tabularx}{\textwidth}{>{\raggedright\arraybackslash}p{3.1cm}
                              >{\raggedright\arraybackslash}p{4.6cm}
                              >{\raggedright\arraybackslash}p{3.0cm}
                              >{\raggedright\arraybackslash}X}
\toprule
\textbf{Newmark CSI Category} & \textbf{Category from \citet{liu2025culturally}} & \textbf{XCR-Bench Category} & \textbf{CSI Example} \\
\midrule

Material Culture &  &  &  \\

\midrule
\multirow{4}{*}{\makecell[l]{Social\\Culture}} 
& Social.Relationship (e.g., Family, Fictive, etc.) 
& Social Etiquette 
& He \textless CSI\textgreater{} thanked \textless/CSI\textgreater{} the stranger who held the elevator door open for him. \\

& Social.Context (e.g., Situational, Historical, etc.) 
& Sports Tradition 
& The family attends \textless CSI\textgreater{} Little League games \textless/CSI\textgreater{} every Saturday during baseball season. \\

& Social.Communicative Goals (e.g., Greeting, Requesting, etc.) 
& Workplace Culture 
& At networking events, attendees often exchange \textless CSI\textgreater{} business cards \textless/CSI\textgreater{} to facilitate future connections. \\

& Social.Demographics (e.g., Income, Education, etc.) 
&  &  \\

\midrule
\multirow{5}{*}{\makecell[l]{Organisations,\\Customs and\\Ideas}} 
& Ideational.Concepts (e.g., Food, Holidays, etc.) 
& Social Tradition 
& They're hosting a \textless CSI\textgreater{} Hogmanay \textless/CSI\textgreater{} celebration to welcome the New Year. \\

& Ideational.Knowledge (e.g., Factual, Common Sense, etc.) 
& Cultural Reference 
& He learned to use \textless CSI\textgreater{} Fahrenheit \textless/CSI\textgreater{} instead of Celsius for temperature discussions. \\

& Ideational.Values (e.g., Bias, Hate, etc.) 
& Cultural Value 
& She donated to a group working to protect \textless CSI\textgreater{} wild horses \textless/CSI\textgreater{}, seeing them as symbols of America's untamed spirit. \\

& Ideational.Norms (e.g., Descriptive, Normative, etc.) 
& Cultural Stereotype 
& They're \textless CSI\textgreater{} helicopter parents \textless/CSI\textgreater{} who monitor their child's every move and decision. \\

& Ideational.artefacts (Art, Meme, Architecture, etc.) 
&  & Walking through the small town, she admired the charming \textless CSI\textgreater{} wooden houses \textless/CSI\textgreater{} that lined the streets. \\

\midrule
Ecology &  &  &  \\

\midrule
Gestures and Habits &  &  &  \\

\bottomrule
\end{tabularx}
\caption{Mapping between Newmark CSI categories, the taxonomy of \citet{liu2025culturally}, and XCR-Bench categories, with illustrative CSI-tagged examples. Empty cells indicate mappings not specified in the current dataset.}
\label{tab:csi_category_mapping}
\end{table*}

\subsection{Adaptation Procedure}
\label{appendix:adaptation_procedure_newmark}

\citet{newmark1988textbook} proposes a set of strategies for translating culture-specific terms, conditioned on contextual and audience-related factors. In this work, we reinterpret these strategies as \emph{cultural adaptation methods} and explicitly prompt models to indicate which method is employed during adaptation. This allows us to analyze which adaptation strategies are most effective for handling CSIs across cultural contexts.

\begin{enumerate}[leftmargin=*, itemsep=2pt]
    \item \textbf{Transference.}  
    The original CSI is retained unchanged in the adapted output (e.g., \emph{sari}, \emph{kimono}).

    \item \textbf{Cultural Equivalent.}  
    A culturally analogous term from the target culture is substituted to convey a similar function or social meaning (e.g., adapting \emph{Thanksgiving} as a local harvest festival).

    \item \textbf{Neutralization.}  
    The CSI is replaced with a description that explains its function or meaning in culturally neutral terms (e.g., rendering \emph{kimono} as ``a traditional Japanese robe'').

    \item \textbf{Literal Translation.}  
    The CSI is translated directly on a word-by-word basis into the target language (e.g., translating \emph{Bundestag} as ``Federal Parliament'').

    \item \textbf{Labeling.}  
    The CSI is retained or adapted and accompanied by a brief explanatory label that clarifies its cultural role or category.

    \item \textbf{Naturalization.}  
    The CSI is adapted to conform to the spelling or pronunciation conventions of the target language (e.g., \emph{Pharisees}).

    \item \textbf{Componential Analysis.}  
    The CSI is decomposed into its constituent semantic components, each of which is explicitly explained (e.g., rendering \emph{dacha} as ``a summer house for wealthy people'').

    \item \textbf{Deletion.}  
    The CSI is omitted entirely when it is non-essential to the meaning or when adaptation would introduce unnecessary complexity.

    \item \textbf{Couplet.}  
    Two adaptation strategies are combined, such as retaining the original CSI while also providing an explanatory description.

    \item \textbf{Accepted Standard Translation.}  
    A widely recognized and conventionally accepted translation is used (e.g., \emph{Holy See} for \emph{Saint-Si\`ege}).

    \item \textbf{Paraphrase or Gloss.}  
    A longer explanatory paraphrase or gloss is added, either inline or as a footnote, to clarify the CSI's meaning and cultural context.

    \item \textbf{Classifier.}  
    A general category term is added to situate the CSI within a familiar conceptual class (e.g., ``the Japanese game of Go'').
\end{enumerate}

\begin{figure}[ht]
  \includegraphics[width=\columnwidth]{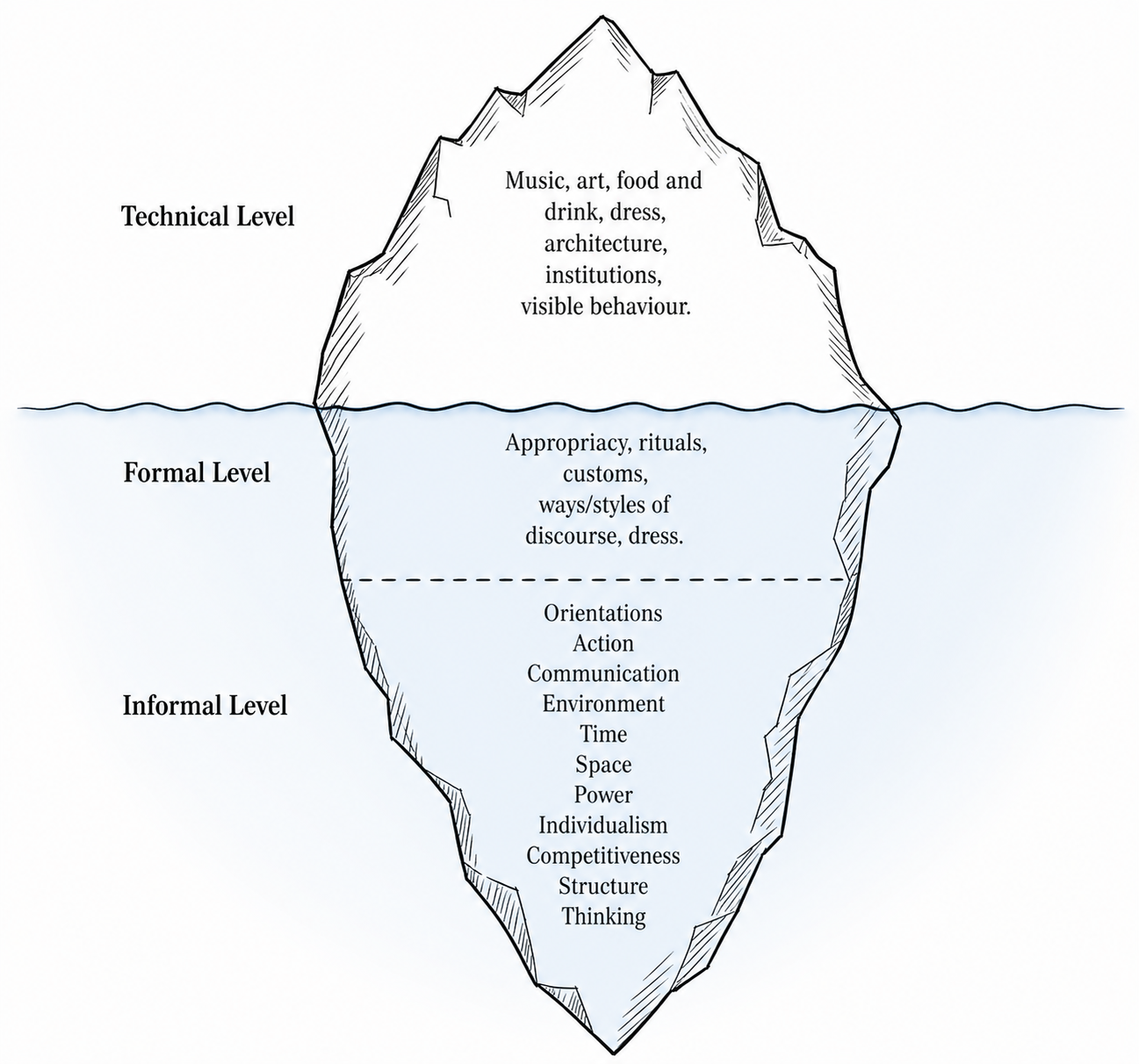}
  \caption{A depiction of Hall's Triad of culture, illustrating three hierarchical levels: Technical (visible artefacts and behaviours), Formal (social norms, rituals, and conventions), and Informal (underlying orientations and value systems). The waterline intersects the Formal level, indicating the boundary between observable and implicit cultural structures.}
  \label{fig:hall_iceberg}
\end{figure}

\section{Hall's Triad of Culture}

In Beyond Culture (1976), Edward T. Hall argued that culture operates largely through unconscious patterns that structure meaning, perception, time, space, and interaction \citep{hall1976beyond}. His distinction between observable practices and the underlying value systems that shape them provides the conceptual basis from which later intercultural scholars derived the now-popular "iceberg" or "triad" metaphor of culture. The metaphor draws on a striking property of icebergs: only about 10\% of an iceberg is visible above the waterline at any given time, while the remaining 90\% lies hidden beneath the surface. By analogy, the visible aspects of culture — language, dress, food, festivals, art — represent only a small fraction of what constitutes a cultural system, with the larger portion comprising values, beliefs, norms, and assumptions that operate beneath conscious awareness.
The conceptual foundations of this metaphor can be traced to two of Hall's earlier works, The Silent Language \citep{hall1973silent} and The Hidden Dimension \citep{hall1966hidden}, which introduced related constructs such as high- and low-context communication, proxemics, and polychronic versus monochronic time orientations. Together, these works established a framework for understanding culture as a multi-layered system in which surface behaviours are governed by deeper, often invisible, structures of meaning.
Despite its widespread influence, the iceberg analogy has notable limitations. As depicted in Figure~\ref{fig:hall_iceberg}, an iceberg is a static and fixed object, whereas culture is inherently dynamic, evolving across generations and reshaped through contact between communities. The analogy is therefore most useful for illustrating the layered structure of culture: its visible and invisible components, rather than its characteristic processes, such as transmission, contestation, and change. In this work, we adopt the iceberg metaphor in this restricted sense: as a heuristic for organising cultural elements by depth of visibility, while recognising that the underlying cultural system it represents is fluid rather than fixed.

\section{Necessity of Integrating Newmark's CSI taxonomy with Hall's Triad}

Newmark's CSI taxonomy and Hall's cultural triad capture orthogonal dimensions of culture: CSI identifies what cultural element is present, while Hall's framework characterizes how deeply its meaning is embedded (visible vs. implicit norms). CSI categories alone do not distinguish surface artefacts from behavioral conventions or underlying beliefs. Hall's stratification enables analysis of whether LLM failures stem from (1) lexical gaps (visible CSIs), (2) pragmatic misunderstanding, or (3) reasoning deficits (semi-visible/invisible CSIs). Thus, Hall's triad provides independent explanatory value by assessing cultural reasoning depth rather than category coverage alone. Prior work, such as \citet{singh2024translating} recognizes the need for Hall's framework in LLM cultural analysis but lacks the granular annotation required for systematic application. Our corpus and benchmark address this gap by operationalizing Hall's depth model within CSI-based evaluation.

\section{Annotation and Annotator Details}
\label{appendix:annotation_procedure}

\textbf{Annotation Training.}
The annotation process was preceded by structured training sessions conducted by the authors and the principal investigator (PI). For each language, we organized two online meetings (approximately one hour each) with both annotators and expert annotators. These sessions focused on clarifying the annotation objectives, discussing the notion of Culture-Specific Items (CSIs), and walking through the detailed annotation guidelines provided in Appendix~\ref{appendix:annotation_guideline}. Annotators were encouraged to ask questions and resolve ambiguities before beginning the annotation process.

\textbf{Annotator Qualifications.}
Annotators were recruited through established academic collaborators in Bengali, Chinese, and Arabic. For each language, the annotation team consisted of two annotators and one expert annotator. Expert annotators hold a PhD in Computational Linguistics or a closely related field and have substantial prior experience in corpus construction and linguistic annotation. The two annotators are either undergraduate or PhD students with prior experience in linguistic or NLP-related annotation tasks.

Specifically, the Arabic annotators are undergraduate students based in Saudi Arabia, the Chinese annotators are PhD students based in the UK, and the Bengali annotators are PhD students based in the UK or the USA, all working in NLP or closely related fields. The two annotators independently produced both intra-lingual and inter-lingual adaptations of the base sentences. In cases of disagreement, the expert annotator adjudicated and selected the most appropriate adaptation.

\textbf{Compensation.}
Annotators are compensated for their work in accordance with the minimum wage standards of the countries in which they reside.

\subsection{General Annotation Guidelines for Cross-Cultural CSI Adaptation}
\label{appendix:annotation_guideline}

You will work with a dataset of English sentences in which culture-specific items (CSIs) which are predominantly drawn from Western (US/UK) cultural contexts. The CSIs are explicitly marked using \texttt{<CSI>} tags in the dataset samples. Your task is to adapt these CSI-marked terms into culturally appropriate counterparts for \textbf{Arabic/Chinese/Bengali} cultures. For each instance, you will produce:
(i) an \textit{intra-lingual} adaptation in English that reflects the target cultural context, and  
(ii) an \textit{inter-lingual} adaptation in your own language.

Please ensure that adaptations maintain the original sentence's communicative purpose while being culturally appropriate and natural in the target context.

To maintain consistency across adaptations, kindly follow the guidelines provided below.

\subsubsection*{1. Purpose}
The goal of annotation is to ensure that CSI adaptations:
\begin{itemize}[itemsep=-2pt, topsep=2pt]
    \item Reflect authentic cultural values, norms, and practices of the target culture.
    \item Use natural and idiomatic language in both intra- and inter-lingual adaptations.
    \item Respect cultural, religious, and social sensitivities, including regional variation where applicable.
\end{itemize}

\subsubsection*{2. Adaptation Rules}

Following Newmark's cultural equivalence framework \citep{newmark1988textbook}, annotate each CSI using one of the following strategies. [For illustration, adaptation examples are given here in Arabic.]

\paragraph{A. Direct Equivalent}
\textbf{Rule:} Replace the CSI with a culturally identical or near-identical equivalent that exists in the target culture.

\textit{Example:}  
Original: \texttt{<CSI>pub gathering</CSI>}  
Intra-lingual (Arabic culture): \texttt{qahwa gathering}  
Inter-lingual (Arabic): \foreignlanguage{arabic}{جلسة قهوة }

\textbf{Annotation tip:} Ensure regional appropriateness (e.g., differences across Arab regions or between Bengali varieties).

\paragraph{B. Functionally Similar Equivalent}
\textbf{Rule:} When no direct equivalent exists, replace the CSI with a culturally different but functionally similar concept.

\textit{Example:}  
Original: \texttt{<CSI>trick-or-treating</CSI>}  
Intra-lingual (Arabic culture): \texttt{children receiving Eid gifts}  
Inter-lingual (Arabic): \foreignlanguage{arabic}{عيدية الأطفال في العيد}

\paragraph{C. Neutral Term}
\textbf{Rule:} If no clear cultural equivalent exists, use a neutral expression that fits social norms in the target culture without introducing foreign practices.

\textit{Example:}  
Original: \texttt{<CSI>friendly banter</CSI>}  
Intra-lingual (Arabic culture): \texttt{light-hearted teasing}  
Inter-lingual (Arabic): \foreignlanguage{arabic}{ مزاحًا خفيفً}

\paragraph{D. Non-transferable}
\textbf{Rule:} Label the CSI as \textit{Non-transferable} if adapting it would conflict with cultural, religious, or social norms in the target culture.

\textit{Example:}  
Original: \texttt{They met through a <CSI>dating app</CSI>.}  
Adaptation: \texttt{Non-transferable}

\subsubsection*{3. Cultural Sensitivity}
While doing the annotation, please ensure to:
\begin{itemize}[itemsep=-2pt, topsep=2pt]
    \item Avoid culturally inappropriate or taboo content (e.g., alcohol, gambling, or sensitive relationships where applicable).
    \item Respect religious norms and social conventions.
\end{itemize}

\subsubsection*{4. Annotation Checklist}
For each sentence, please verify:
\begin{itemize}[itemsep=-2pt, topsep=2pt]
    \item \textbf{Cultural accuracy:} Is the CSI adapted appropriately for the target culture?
    \item \textbf{Linguistic quality:} Is the adaptation natural and idiomatic?
    \item \textbf{Functional equivalence:} Does the adaptation preserve the original communicative intent?
    \item \textbf{Consistency:} Do the intra- and inter-lingual adaptations align?
\end{itemize}

\subsubsection*{5. Final Notes}
\begin{itemize}[itemsep=-2pt, topsep=2pt]
    \item Consult the expert annotator for ambiguous or borderline cases.
    \item Flag creative or uncertain adaptations for further review.
    \item Use {\Large C}{\normalsize ANDLE} and Cultural Atlas as reference sources for relevant cultural practices, rituals, and traditions.
\end{itemize}

\begin{figure*}[ht]
  \includegraphics[width=\textwidth]{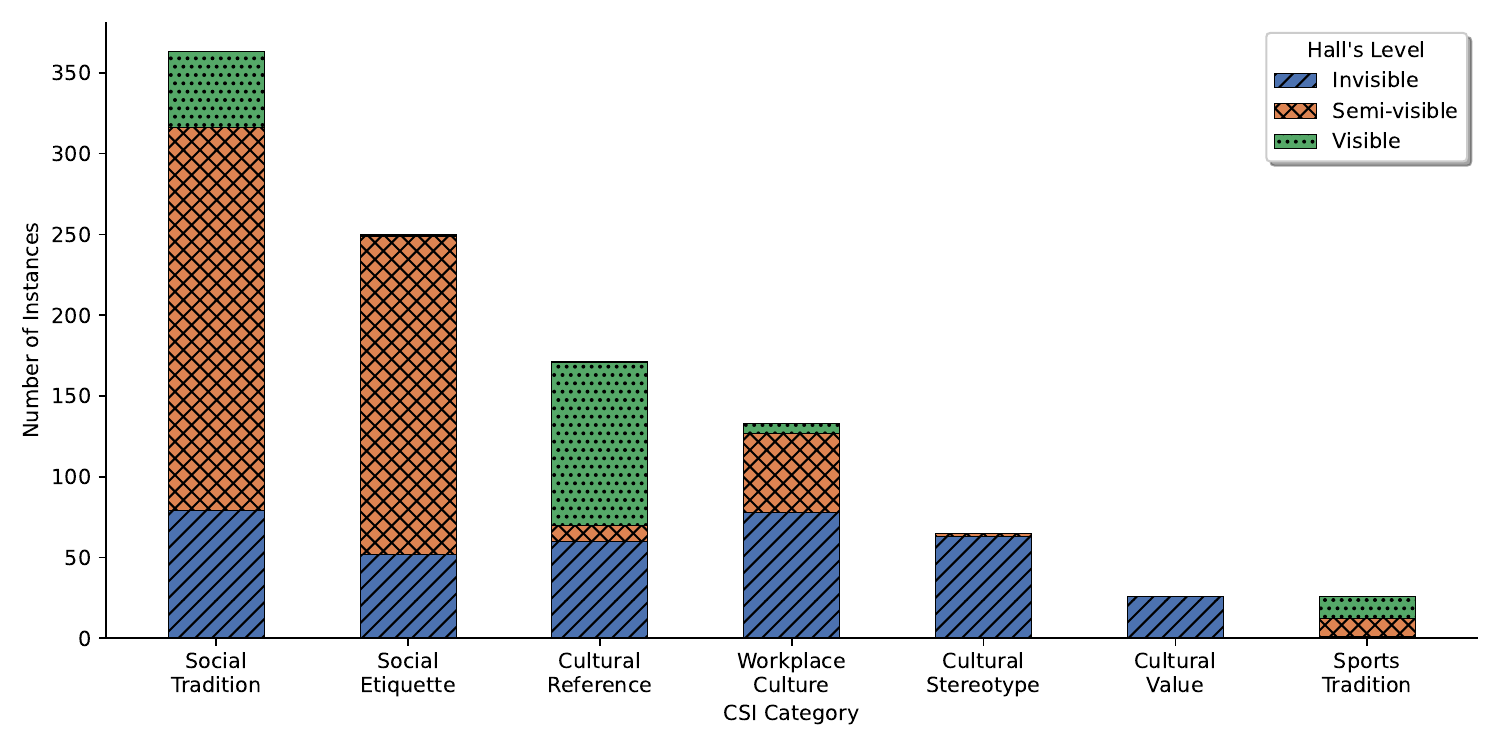}
  \caption{Mapping between CSI categories and Hall's cultural levels. The textured bar plots indicate a higher concentration of semi-visible and invisible cultural elements across all CSI categories.}
  \label{fig:csi_hall}
\end{figure*}









\begin{table*}[ht]
\centering
\small
\begin{tabular}{p{0.25\linewidth} p{0.70\linewidth}}
\toprule
\textbf{Field} & \textbf{Content} \\
\midrule

\textbf{CSI} & Halloween \\

\textbf{Cultural Context} &
Halloween is widely celebrated in the West with decorations, costumes, and trick-or-treating. \\

\textbf{Sentence} &
We're decorating the house for \texttt{<CSI>} Halloween \texttt{</CSI>} this weekend. \\

\textbf{CSI Category} & Social Tradition \\

\textbf{CSI-Hall Mapping} & Rituals \\

\textbf{Hall's Level} & Semi-visible \\

\specialrule{0.9pt}{0pt}{0pt}
\multicolumn{2}{l}{\textbf{Chinese}} \\
\midrule

\textit{Intra-lingual} &
We're decorating the house for the \texttt{<CSI>} Lantern Festival \texttt{</CSI>} this weekend with lanterns and festive lights. \\

\textit{Inter-lingual} & 

\foreignlanguage{chinese}{这个周末我们正在为\texttt{<CSI>}元宵节\texttt{</CSI>}装饰房子，挂上彩灯，营造节日的夜晚氛围。}\\

\specialrule{0.9pt}{0pt}{0pt}
\multicolumn{2}{l}{\textbf{Arabic}} \\
\midrule

\textit{Intra-lingual} &
We're decorating the house for \texttt{<CSI>} Eid celebration \texttt{</CSI>} this weekend. \\

\textit{Inter-lingual} &
\foreignlanguage{arabic}{نقوم بتزيين المنزل لـ}  \texttt{</CSI>}\foreignlanguage{arabic}{ احتفال العيد  } \texttt{</CSI>}\foreignlanguage{arabic}{ هذا الأسبوع.} \\

\specialrule{0.9pt}{0pt}{0pt}
\multicolumn{2}{l}{\textbf{Bengali (West Bengal)}} \\
\midrule

\textit{Intra-lingual} &
We're decorating the house for \texttt{<CSI>} Durga Puja \texttt{</CSI>} this weekend. \\

\textit{Inter-lingual} &
\foreignlanguage{bengali}{আমরা এই সপ্তাহান্তে বাড়ি \texttt{<CSI>}দুর্গা পূজার\texttt{</CSI>} জন্য সাজাচ্ছি।} \\

\specialrule{0.9pt}{0pt}{0pt}
\multicolumn{2}{l}{\textbf{Bengali (Bangladesh)}} \\
\midrule

\textit{Intra-lingual} &
We're decorating the house for \texttt{<CSI>} Eid-ul-Fitr \texttt{</CSI>} this weekend. \\

\textit{Inter-lingual} &
\foreignlanguage{bengali}{আমরা এই সপ্তাহান্তে বাড়ি \texttt{<CSI>}ঈদুল ফিতরের\texttt{</CSI>} জন্য সাজাচ্ছি।} \\

\bottomrule
\end{tabular}
\caption{An example sample from the \textsc{XCR-Bench} corpus along with the parallel Inter-lingual and Intra-lingual annotations in other cultures.}
\label{tab:csi_halloween_example}
\end{table*}

\section{Properties of \textsc{XCR-Bench} Corpus}
\label{appendix:dataset_properties}

\subsection{Data Acquisition}
{\Large C}{\normalsize ANDLE} and Cultural Atlas are used as the primary knowledge bases for data acquisition to construct the \textsc{XCR-Bench} base corpus.
{\Large C}{\normalsize ANDLE} provides high-quality cultural commonsense knowledge (CCSK), comprising $1.1$M assertions organized into $60$K coherent clusters, while Cultural Atlas offers curated descriptions of multicultural norms derived from extensive global community interviews and rigorous validation \citep{rao2024normad}. Both resources present cultural knowledge in factual form (e.g., {\Large C}{\normalsize ANDLE} states a Representative cultural concept as \textit{``In the USA, people wear their wedding rings on the ring finger of their left hand''}), with Cultural Atlas further organizing information into thematic sections such as Core Concepts, Communication, and Business Culture.

\subsection{Dataset Properties}
As detailed in Section \ref{sec:dataset_analysis}, \textsc{XCR-Bench} is the first dataset to provide comprehensive annotations aligned with Hall's triad of culture, with particular emphasis on the \textit{Semi-visible} and \textit{Invisible} layers that encompass beliefs, values, and social norms. Figure~\ref{fig:csi_hall} illustrates the mapping between CSI categories and Hall's cultural levels, highlighting a predominant focus on these \textit{Semi-visible} and \textit{Invisible} layers.

Table \ref{tab:csi_halloween_example} shows an example sample from the \textsc{XCR-Bench} corpus. Each sample contains its fine-grained annotation for CSI Category, CSI-Hall Mapping, Hall's Level and parallel inter-lingual and intra-lingual cross-cultural CSI pairs in four distinct cultures. 

\begin{figure*}[ht]
  \includegraphics[width=\textwidth]{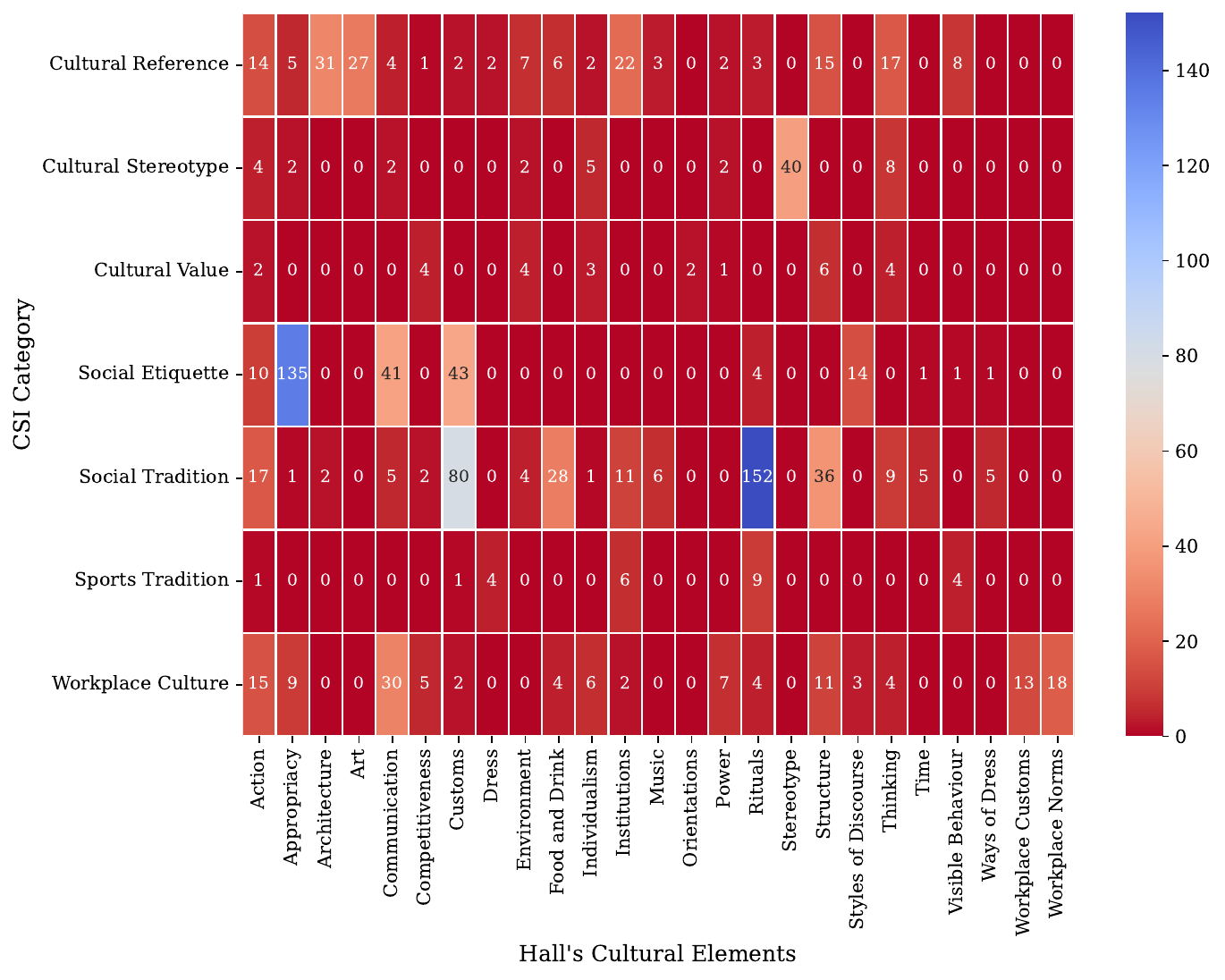}
  \caption{Mapping between CSI categories and Hall's cultural elements in the XCR-Bench Corpus.}
  \label{fig:csi_hall_elements}
\end{figure*}

We present the mapping of Hall's cultural elements with our CSI categories in Figure \ref{fig:csi_hall_elements}. This shows how the Hall's cultural elements are mapped across our $7$ distinct categories of CSIs. Some CSI categories and cultural elements are predominant in number because of their vast availability in the source where we extracted our CSIs from, namely {\Large C}{\normalsize ANDLE}
 and Cultural Atlas.

We explicitly show the cultural element mapping with Hall's triads in Table \ref{tab:hall_visibility_mapping} with instances from our \textsc{XCR-Bench} corpus. We derive this mapping from \citet{hall1976beyond}. Notably, the Hall cultural level of CSIs is independent of Newmark's categorization and based on their context.

\begin{table*}[ht]
\centering
\small
\begin{tabular}{p{0.20\linewidth} p{0.20\linewidth}p{0.60\linewidth}}
\toprule
\textbf{Hall's Level} & \textbf{Cultural Element} & \textbf{Instance from \textsc{XCR-Bench}} \\
\midrule

\multirow{10}{*}{\textbf{Invisible}} 
& Action & He's an <CSI> adrenaline junkie </CSI> who loves skydiving and bungee jumping. \\
& Communication & He steered conversation away from <CSI> gun laws </CSI> to maintain pleasant dinner atmosphere.\\
& Competitiveness & He’s a <CSI> self-made </CSI> man who built his empire from scratch. \\\\
& Environment & She loves going to the <CSI> pub </CSI> for a pint after work. \\ \\
& Individualism & She demonstrates <CSI> self-reliance </CSI> by managing projects independently.\\
& Power & The board member <CSI> endorses a bottom-up approach </CSI>, empowering employees to contribute to decision-making. \\
& Stereotype & He's the <CSI> all-American boy next door </CSI> with his football scholarship and volunteer work. \\
& Structure & Couples are <CSI> delaying parenthood </CSI> to establish <CSI> financial stability </CSI> first. \\
& Thinking & The building omitted floor <CSI> 13 </CSI> jumping from 12 to 14 due to <CSI> superstition </CSI>. \\
& Time & He made sure to arrive 10 minutes early for the meeting because <CSI> punctuality </CSI> is important in American culture. \\

\midrule

\multirow{8}{*}{\textbf{Semi-visible}} 
& Appropriacy & Don't ask about <CSI> someone's weight or appearance </CSI> in personal conversations. \\
& Customs & She is used to <CSI> queue </CSI> politely and wait her turn, even in crowded places. \\
& Rituals & She always says <CSI> grace </CSI> before dinner, a tradition passed down through generations. \\
& Styles of Discourse & She said <CSI> 'cheers' </CSI> when the stranger held the door, a casual British thanks. \\
& Behavior & He crossed his <CSI> index and middle fingers </CSI> for <CSI> good luck </CSI> before the exam. \\
& Ways of Dress & He wore a tailored <CSI> suit </CSI> to the job interview but dressed in casual <CSI> jeans </CSI> and a t-shirt for the barbecue.\\
& Workplace Customs & We're implementing an <CSI> open-door policy </CSI> in management for better accessibility and transparency. \\
& Workplace Norms & The manager gave <CSI> constructive criticism </CSI> during the performance review. \\

\midrule

\multirow{6}{*}{\textbf{Visible}} 
& Architecture & Their dining table, made of <CSI> White Oak </CSI>, has been in the family for generations, a testament to its durability. \\
& Art & When she visited the UK, she noticed that every British coin had a portrait of the reigning <CSI> monarch </CSI>, a tradition that felt like holding a piece of history. \\
& Dress & Everyone wore their team <CSI> merchandise </CSI> on game day to show support. \\
& Food and Drink & They enjoyed a <CSI> fish and chips supper </CSI> by the seaside, a classic British experience. \\
& Institutions & The startup has an <CSI> open office layout </CSI> without walls or dividers. \\\\
& Music & Listening to a British <CSI> brass band </CSI> at the local fair, she was struck by the rich, traditional sound. \\

\bottomrule
\end{tabular}
\caption{Mapping of cultural elements to Hall's Triad of Culture with instances from our constructed \textsc{XCR-Bench} Corpus. The mapping is derived from \citet{hall1976beyond}.}
\label{tab:hall_visibility_mapping}
\end{table*}




\newpage
\onecolumn

\section{Details on Evaluation Metrics}
\label{appendix:detailed_evaluation_metrics}

A summary of the evaluation metrics used in this study is presented in Table \ref{tab:eval-metrics-details}. 

\begin{table}[!h]
\centering
\resizebox{\textwidth}{!}{
\begin{tabular}{@{}l l p{6cm} p{6cm}@{}}
\toprule
\textbf{Task} & \textbf{Metric} & \textbf{Description} & \textbf{Formula/Details} \\
\midrule
\multirow{2}{*}{\textbf{CSI Identification}} & Hard Identification of CSI (\textbf{HI-CSI}) & Strict matching using exact string equality. & $\mathrm{sim}_{\text{exact}}(g,m)=\mathbb{I}[g=m]$, where $g$ and $m$ denote gold and model-predicted CSI spans. \\
\cline{2-4}
& Soft Identification of CSI (\textbf{SI-CSI}) & Accounts for minor variations via normalized Levenshtein similarity, with optimal alignment using the Hungarian algorithm. & $s(g,m)=1-\frac{d_{\text{Lev}}(g,m)}{\max(|g|,|m|)}$; sentence-level soft F$_1$ based on alignment. \\
\midrule
\multirow{2}{*}{\textbf{CSI Prediction}} & Hard Prediction of CSI (\textbf{HP-CSI}) & Case-insensitive exact match between gold and predicted spans. Span refers to text inside the enclosed \texttt{<CSI>} and \texttt{</CSI>} tags. & $\mathrm{sim}_{\text{exact}}(g,p)=\mathbb{I}[\mathrm{lower}(g)=\mathrm{lower}(p)]$, where $g$ and $p$ denote gold and predicted spans. \\
\cline{2-4}
& Soft Prediction of CSI (\textbf{SP-CSI}) & Cosine similarity between Sentence-BERT embeddings for paraphrases/synonyms. & $\mathrm{sim}_{\text{sem}}(g,p) = \cos\big(\mathbf{e}(g), \mathbf{e}(p)\big)$, using \texttt{all-mpnet-base-v2} \citep{song2020mpnet}. \\
\midrule 
\multirow{3}{*}{\textbf{CSI Adaptation}} & CSI-Level Semantic Similarity (\textbf{CSI$_\text{bert}$}) & CSI-level BERTScore F1 for aligned CSI spans. & $\mathrm{CSI}_{\text{bert}}=\frac{1}{n}\sum_{i=1}^{n}\mathrm{F1}_{\text{bert}}(g_i,o_i)$, where $\mathcal{G}=\{g_i\}$ and $\mathcal{O}=\{o_i\}$ are gold and predicted CSI spans extracted from \texttt{<CSI>} tags, respectively.\\
\cline{2-4}
&  Sentence-Level Semantic Similarity (\textbf{SENT$_\text{bert}$}) & Assesses sentence-level meaning preservation using BERTScore F1. & $\mathrm{SENT}_{\text{bert}} = \mathrm{F1}_{\text{bert}}\big(s^{\text{out}}, s^{\text{gt}}\big)$, where $s^{\text{out}}$ and $s^{\text{gt}}$ are model-generated and ground-truth sentences. \\
\bottomrule
\end{tabular}
}
\caption{Evaluation metrics for the three cross-cultural reasoning tasks: CSI Identification, Prediction, and Adaptation. Hard metrics enforce strict correctness, while soft metrics provide graded credit for semantically or superficially similar outputs.}
\label{tab:eval-metrics-details}
\end{table}

\subsection{CSI Identification}

We evaluate CSI identification using two complementary metrics: a strict exact-match score and a soft similarity-based score.\\\\
\textbf{Hard Identification of CSI (HI-CSI).}
HI-CSI enforces strict exact matching, rewarding only predictions that precisely replicate ground-truth CSIs. This metric is particularly suited for evaluating models in high-stakes applications where partial identifications may not suffice, such as in content moderation systems requiring unambiguous detection.
The sentence-level score $\text{HI-CSI}$ is defined as follows:
\begin{itemize}
\item If $|G_s| = 0$:
$$\text{HI-CSI} = 
\begin{cases} 
1 & \text{if } |P_s| = 0, \\
0 & \text{otherwise}.
\end{cases}$$
This penalizes false-positive predictions when no CSIs are present in the ground truth.
\item If $|G_s| > 0$:
$$\text{HI-CSI}_s = \frac{1}{|G_s|} \sum_{g \in G_s} \mathbb{I}(g \in P_s),$$

where $\mathbb{I}(\cdot)$ is the indicator function, equal to 1 if the condition holds (i.e., an exact string match exists in $P_s$) and 0 otherwise. Note that if duplicate CSIs appear in $G_s$, each is evaluated independently, though matches are checked via membership, allowing a single matching prediction to satisfy multiple identical ground-truth entries if present.
\end{itemize}
The overall HI-CSI score for the dataset is the average across all sentences, scaled to a percentage for interpretability:
$$\text{HI-CSI} = \left( \frac{1}{|\mathcal{D}|} \sum_{s \in \mathcal{D}} \text{HI-CSI}_s \right)$$
This ranges from 0 (no matches) to 1 (perfect alignment across all sentences).\\ \\
\textbf{Soft Identification of CSI (SI-CSI).}To account for minor surface-form variations, we compute a soft CSI matching score based on normalized Levenshtein similarity.
For each sentence $n$, the extracted ground-truth CSI set
$G^{(n)}=\{g^{(n)}_1,\dots,g^{(n)}_{m_n}\}$ and the predicted CSI set
$P^{(n)}=\{p^{(n)}_1,\dots,p^{(n)}_{k_n}\}$ are compared using a similarity-based
one-to-one matching procedure, where $m_n=|G^{(n)}|$ and $k_n=|P^{(n)}|$
denote the numbers of ground-truth and predicted CSIs, respectively.
 
Pairwise normalised Levenshtein similarities $s(g^{(n)}_i,p^{(n)}_j)=1-\frac{d_L(g^{(n)}_i,p^{(n)}_j)}{\max(\ell(g^{(n)}_i),\ell(p^{(n)}_j))}$ are first computed between all ground-truth and predicted CSIs, and an optimal one-to-one alignment is then obtained using the Hungarian algorithm so that the total similarity of the matched pairs is maximized while preventing reuse of any CSI. Based on the optimal one-to-one alignment, a soft F1 score is computed for each sentence as
\[
\text{SI-CSI}
=
\mathrm{\text{Soft (F1)}}_n
=
\mathbf{1}[m_n=0 \land k_n=0]
+
\mathbf{1}[m_n>0 \land k_n>0]\;
\frac{2}{m_n+k_n}
\sum_{i=1}^{m_n}\sum_{j=1}^{k_n}
s(g^{(n)}_i,p^{(n)}_j)\,X^{*(n)}_{ij},
\]

where $X^{*(n)}_{ij}$ denotes the optimal one-to-one assignment variable, which takes the value $1$ when $g^{(n)}_i$ is matched to $p^{(n)}_j$ and $0$ otherwise, so that both missing ground-truth CSIs and over-generated predictions are penalized.\\

HI-CSI measures strict span identification accuracy, while SI-CSI provides graded credit for near-matches without relying on heuristic substring rules.

\paragraph{On the relative scores of HI-CSI and SI-CSI.}
A natural expectation might be that the soft metric (SI-CSI) should always score higher than the hard metric (HI-CSI), since soft matching is more lenient. However, our results show the opposite for several models. This arises from a fundamental difference in what the two metrics measure. HI-CSI is a recall-only metric that checks whether each gold CSI span is found in the model's predictions; extra (false-positive) predictions are ignored. SI-CSI, in contrast, computes an F1 score using normalised Levenshtein similarity with optimal one-to-one alignment, and therefore penalises both missed gold spans (recall) and unjustified extra predictions (precision).

As a result, over-prediction lowers SI-CSI without affecting HI-CSI. For example, given the gold span ``Christmas dinner'' and a model prediction ``Christmas dinner, turkey, stuffing'', HI-CSI scores $1.0$ (the gold span is recovered exactly), while SI-CSI's precision drops to $1/3$ due to the two unjustified extra spans, lowering the F1 score correspondingly. This pattern is most pronounced for models with high recall but low precision (e.g., Llama-3.3 in Table~\ref{tab:csi_west_results}), which tend to over-tag spans as CSIs. The two metrics are therefore complementary: HI-CSI measures whether the model can locate culturally salient items at all, while SI-CSI additionally penalises models that flag culturally irrelevant content as CSIs.

\subsection{CSI Prediction}

We evaluate CSI prediction using two complementary metrics: a strict exact-match score and a soft semantic similarity score.\\\\
\textbf{Hard Prediction of CSI (HP-CSI).}
HP-CSI requires strict exact matching (case-insensitive) for paired CSIs, rewarding only precise replications in the corresponding order.
The sentence-level score $\text{HP-CSI}_s$ is defined as:
$$\text{HP-CSI}_s = \sum_{i=1}^{\min(n, m)} \mathbb{I}(g_i.\text{lower()} = p_i.\text{lower()}),$$
where $\mathbb{I}(\cdot)$ is the indicator function, equal to 1 if the condition holds and 0 otherwise. This sums the number of exact matches across the paired elements up to the minimum length.
The overall HP-CSI score for the dataset is the average across all sentences, scaled to account for varying CSI counts (interpretable as a percentage when normalized by average CSI count per sentence):
$$\text{HP-CSI} = \left( \frac{1}{|\mathcal{D}|} \sum_{s \in \mathcal{D}} \text{HP-CSI}_s \right)$$
This value scales with the average number of CSIs per sentence; for datasets with one CSI per sentence, it ranges from 0 to 1.\\
\textbf{Soft Prediction of CSI (SP-CSI).} SP-CSI extends HP-CSI by incorporating semantic similarity for non-exact pairs, using cosine similarity on embeddings generated by the SentenceTransformer model (`all-mpnet-base-v2'). This allows partial credit for predictions that capture similar meaning despite lexical differences.
The sentence-level score $\text{SP-CSI}_s$ is defined as:
$$\text{SP-CSI}_s = \sum_{i=1}^{\min(n, m)} m_i,$$
where for each paired index $i$,
$$m_i = 
\begin{cases} 
1 & \text{if } g_i.\text{lower()} = p_i.\text{lower()} \text{ (exact match)}, \\
\cos(\mathbf{e}(g_i), \mathbf{e}(p_i)) & \text{otherwise},
\end{cases}$$
and $\mathbf{e}(\cdot)$ denotes the embedding function from the SentenceTransformer model, with $\cos(\mathbf{u}, \mathbf{v}) = \frac{\mathbf{u} \cdot \mathbf{v}}{||\mathbf{u}|| \, ||\mathbf{v}||}$ being the cosine similarity.
The overall SP-CSI score is:
$$\text{SP-CSI} = \left( \frac{1}{|\mathcal{D}|} \sum_{s \in \mathcal{D}} \text{SP-CSI}_s \right)$$\\
This value scales with the average number of CSIs per sentence.

\subsection{CSI Adaptation}
We evaluate CSI adaptation along two complementary dimensions: (i) semantic fidelity of the adapted CSIs themselves, and (ii) overall sentence-level semantic preservation, as sentence structure may change during adaptation. We perform both intra-lingual (English-to-English) and inter-lingual (cross-language) evaluations.
\paragraph{Sentence-Level Semantic Similarity ($\mathrm{SENT}_{\text{bert}}$).}
To assess whether the adapted sentence preserves the meaning of the original, we compute BERTScore F$_1$ between the model-generated sentence $s^{\text{out}}$ and the ground-truth adapted sentence $s^{\text{gt}}$:
$$\mathrm{SENT}_{\text{bert}} = \mathrm{F1}_{\text{bert}}\big(s^{\text{out}}, s^{\text{gt}}\big).$$
BERTScore leverages contextual embeddings from BERT to compute F1 scores based on pairwise token similarities, making it particularly effective for assessing paraphrase quality and semantic preservation in natural language generation tasks. With baseline rescaling enabled to normalize scores relative to a random baseline, improving interpretability and cross-model comparability, the sentence-level full-sentence score is $\text{SENT}_{\text{bert}, s} = \text{F1}'_{\text{BERT}}(a_s, g_s)$, using the entire strings $a_s$ and $g_s$ (where $g_s$ is the ground-truth sentence and $a_s$ the adapted output). The overall dataset-level metric is the average: $\text{SENT}_{\text{bert}} = \frac{1}{|\mathcal{D}|} \sum_{s \in \mathcal{D}} \text{SENT}_{\text{bert}, s}$, where $\mathcal{D}$ denotes the dataset of sentence pairs.
\paragraph{CSI-Level Semantic Similarity ($\mathrm{CSI}_{\text{bert}}$).}
Given aligned gold and predicted CSI spans $\mathcal{G}=\{g_i\}_{i=1}^{n}$ and $\mathcal{O}=\{o_i\}_{i=1}^{n}$ extracted from \texttt{<CSI>} tags, we compute BERTScore F$_1$ for each pair:
$$\mathrm{sim}_{\text{CSI}}(g_i,o_i)=\mathrm{F1}_{\text{bert}}(g_i,o_i).$$
The final CSI-level score is the average across all spans:
$$\mathrm{CSI}_{\text{bert}}=\frac{1}{n}\sum_{i=1}^{n}\mathrm{sim}_{\text{CSI}}(g_i,o_i).$$
CSIs are extracted from both ground-truth and adapted texts via regular expressions matching content within \texttt{<CSI>} tags, producing ordered lists of strings. We enforce that the number of extracted CSIs matches between ground-truth and output for each sentence, ensuring paired evaluation. Let $G_s = [g^{(s)}_1, g^{(s)}_2, \dots, g^{(s)}_{k_s}]$ be the ordered list of ground-truth CSIs (where $k_s = |G_s|$) and $A_s = [a^{(s)}_1, a^{(s)}_2, \dots, a^{(s)}_{k_s}]$ the corresponding list of adapted CSIs, with $|G_s| = |A_s| = k_s$. The sentence-level CSI score aggregates per-CSI F1 scores with baseline rescaling: $\text{CSI}_{\text{BERT}, s} = \frac{1}{k_s} \sum_{i=1}^{k_s} \text{F1}'_{\text{BERT}}(a^{(s)}_i, g^{(s)}_i)$, if $k_s > 0$; otherwise, it defaults to 0 (though datasets typically ensure CSIs exist). The overall dataset-level metric is the average: $\text{CSI}_{\text{BERT}} = \frac{1}{|\mathcal{D}|} \sum_{s \in \mathcal{D}} \text{CSI}_{\text{BERT}, s}$. Category-wise scores can be derived by partitioning $\mathcal{D}$ into subsets $\mathcal{D}_c$ based on CSI categories and computing averages within each.
\paragraph{Intra- vs. Inter-Lingual Evaluation.}
For intra-lingual (English) adaptation, we compute BERTScore using \texttt{bert-base-uncased}. For inter-lingual adaptation, we employ language-specific BERTScore models: \texttt{bert-base-arabic} \citep{safaya-etal-2020-kuisail} for Arabic, \texttt{chinese-bert-wwm-ext} \citep{chinese-bert-wwm} for Chinese, and \texttt{bangla-bert-base} \citep{Sagor_2020} for Bengali, to ensure that semantic similarity is evaluated within the appropriate linguistic and cultural representation space. For inter-lingual evaluation, we extend this framework using multilingual models (e.g., \texttt{bert-base-multilingual-uncased}) to handle cross-language pairs, computing analogous $\text{SENT}_{\text{BERT-multi}}$ and $\text{CSI}_{\text{BERT-multi}}$ scores.

\paragraph{Discussion and Rationale.}
These metrics emphasize semantic equivalence, with $\text{CSI}_{\text{BERT}}$ isolating performance on culturally sensitive adaptations- critical for ensuring biases or sensitivities are appropriately mitigated without altering core meaning. The use of baseline rescaling maps scores to a more intuitive [0,1] range, where higher values indicate better alignment. Compared to n-gram-based metrics like BLEU, BERTScore better handles paraphrasing and synonyms, aligning with adaptation goals. Limitations include: (1) reliance on BERT's English-centric biases in intra-lingual settings and (2) computational cost and latency from embedding computations, though batched inference mitigates this. In practice, metrics are implemented using the Hugging Face \texttt{bert-score} library, with time complexity dominated by BERT inference: $O(|\mathcal{D}| \times (L_s^2))$ per sentence for greedy matching in BERTScore, where $L_s$ is token length (typically <512).

\section{Additional Results}
\label{appendix:additional_results}

\subsection{CSI Identification \& Prediction}


\begin{table}[ht]
\centering
\small
\setlength{\tabcolsep}{4pt}
\begin{tabular}{@{}llccc@{\hspace{8pt}}ccc@{}}
\toprule
\multirow{2}{*}{\textbf{Task}} & \multirow{2}{*}{\textbf{Metric}} & 
\multicolumn{3}{c}{\textbf{Mean $\pm$ SE (\%)}} & 
\multicolumn{3}{c}{\textbf{Wilcoxon $p$-value (one-sided)}} \\
\cmidrule(lr){3-5}\cmidrule(lr){6-8}
& & \textbf{V} & \textbf{SV} & \textbf{I} & \textbf{V$>$SV} & \textbf{SV$>$I} & \textbf{V$>$I} \\
\midrule
\multirow{2}{*}{\textbf{Identification}} 
& Hard (HI-CSI) & $58.07 \pm 2.62$ & $48.61 \pm 1.68$ & $51.81 \pm 1.23$ 
  & $0.004^{**}$ & $0.981$ & $0.012^{*}$ \\
& Soft (SI-CSI) & $61.20 \pm 1.27$ & $56.48 \pm 1.43$ & $55.99 \pm 1.62$ 
  & $0.008^{**}$ & $0.473$ & $0.004^{**}$ \\
\midrule
\multirow{2}{*}{\textbf{Prediction}}
& Hard (HP-CSI) & $40.66 \pm 3.08$ & $39.54 \pm 2.35$ & $40.59 \pm 2.46$ 
  & $0.320$ & $0.844$ & $0.527$ \\
& Soft (SP-CSI) & $79.18 \pm 1.79$ & $77.35 \pm 2.05$ & $72.64 \pm 2.30$ 
  & $0.074$ & $0.004^{**}$ & $0.004^{**}$ \\
\bottomrule
\end{tabular}
\caption{Performance across Hall's cultural triad (V = Visible, SV = Semi-visible, I = Invisible) for the CSI Identification and Prediction tasks under hard and soft metrics. Mean $\pm$ standard error reported across 8 LLMs. One-sided Wilcoxon signed-rank tests assess whether higher-visibility levels yield higher scores. Significance: $^{*}p<0.05$, $^{**}p<0.01$, $^{***}p<0.001$. Soft metrics consistently reveal a Visible-vs-Invisible decline; hard metrics show a less monotonic pattern, particularly for Prediction.}
\label{tab:hall_identification_prediction}
\end{table}


\begin{table*}[ht]
\centering
\tiny
\setlength{\tabcolsep}{2.6pt}
\renewcommand{\arraystretch}{1.05}
\begin{adjustbox}{max width=\textwidth}
\begin{tabular}{l*{8}{cc}}
\toprule
\multirow{2}{*}{\textbf{Hall's Cultural Element}} &
\multicolumn{2}{c}{\textbf{DS-Distill}} &
\multicolumn{2}{c}{\textbf{Claude}} &
\multicolumn{2}{c}{\textbf{DS-R1}} &
\multicolumn{2}{c}{\textbf{Gemini}} &
\multicolumn{2}{c}{\textbf{GPT-4o}} &
\multicolumn{2}{c}{\textbf{Llama-3.3}} &
\multicolumn{2}{c}{\textbf{OLMo-2}} &
\multicolumn{2}{c}{\textbf{Qwen-2.5}} \\
\cmidrule(lr){2-3}\cmidrule(lr){4-5}\cmidrule(lr){6-7}\cmidrule(lr){8-9}
\cmidrule(lr){10-11}\cmidrule(lr){12-13}\cmidrule(lr){14-15}\cmidrule(lr){16-17}
& \textbf{HI} & \textbf{SI} & \textbf{HI} & \textbf{SI} & \textbf{HI} & \textbf{SI} & \textbf{HI} & \textbf{SI}
& \textbf{HI} & \textbf{SI} & \textbf{HI} & \textbf{SI} & \textbf{HI} & \textbf{SI} & \textbf{HI} & \textbf{SI} \\
\midrule
Action & 51.43 & 54.91 & 67.14 & 73.51 & 64.29 & 63.74 & 58.57 & 63.64 & 65.71 & 66.35 & 61.43 & 51.43 & 54.29 & 59.68 & 61.43 & 66.04 \\
\textbf{Appropriacy} & \underline{19.40} & \underline{39.86} & \underline{19.40} & \underline{38.02} & \underline{24.63} & \underline{39.32} & \underline{18.66} & \underline{30.85} & \underline{23.88} & \underline{37.76} & \underline{23.13} & \underline{30.31} & \underline{22.39} & \underline{38.73} & \underline{32.84} & \underline{46.38} \\
Architecture & 38.89 & 48.55 & 38.89 & 50.65 & 72.22 & 67.84 & 55.56 & 63.23 & 55.56 & 57.61 & 41.67 & 45.70 & 33.33 & 46.40 & 55.56 & 65.57 \\
Art & 60.00 & 53.25 & 60.00 & 62.84 & 73.33 & 59.34 & 50.00 & 53.64 & 60.00 & 55.95 & 93.33 & 65.37 & 50.00 & 54.46 & 70.00 & 57.14 \\
Communication & 36.49 & 46.60 & 40.54 & 52.31 & 36.49 & 44.81 & 27.03 & 39.54 & 24.32 & 42.66 & 28.38 & 35.94 & 33.78 & 53.13 & 32.43 & 46.81 \\
Competitiveness & 68.75 & 66.23 & 62.50 & 68.65 & 62.50 & 68.63 & 56.25 & 66.99 & 56.25 & 66.99 & 62.50 & 63.80 & 56.25 & 69.04 & 62.50 & 66.96 \\
Customs & 43.22 & 55.94 & 38.14 & 56.83 & 50.85 & 58.50 & 38.98 & 51.66 & 40.68 & 53.54 & 42.37 & 46.72 & 29.66 & 44.61 & 47.46 & 55.98 \\
Dress & 25.00 & 42.42 & 50.00 & 56.10 & 50.00 & 56.10 & 25.00 & 41.51 & 25.00 & 44.91 & 50.00 & 50.93 & 25.00 & 41.44 & 50.00 & 56.94 \\
Environment & 88.89 & 65.43 & 77.78 & 76.94 & 88.89 & 78.40 & 88.89 & 82.41 & 66.67 & 65.24 & 55.56 & 48.87 & 77.78 & 85.89 & 77.78 & 69.88 \\
Food and Drink & 72.22 & 72.32 & 66.67 & 75.28 & 75.00 & 75.34 & 69.44 & 73.80 & 72.22 & 77.20 & 55.56 & 55.47 & 69.44 & 83.09 & 63.89 & 64.71 \\
Individualism & 38.89 & 53.71 & 33.33 & 53.35 & 55.56 & 56.64 & 33.33 & 43.08 & 55.56 & 49.34 & 55.56 & 55.20 & 50.00 & 64.64 & 55.56 & 63.23 \\
Institutions & 71.05 & 61.02 & 50.00 & 60.87 & 74.56 & 61.14 & 51.75 & 56.25 & 64.04 & 61.14 & 78.95 & 59.58 & 41.23 & 60.65 & 71.93 & 63.64 \\
Music & 40.00 & 49.13 & 60.00 & 72.83 & 70.00 & 82.40 & 70.00 & 84.44 & 30.00 & 61.39 & 70.00 & 62.17 & 50.00 & 74.72 & 70.00 & 82.40 \\
\textbf{Orientations} & \underline{0.00} & \underline{20.00} & \underline{0.00} & \underline{20.00} & \underline{100.00} & \underline{33.33} & \underline{100.00} & \underline{66.67} & \underline{0.00} & \underline{7.41} & \underline{100.00} & \underline{33.33} & \underline{0.00} & \underline{0.00} & \underline{0.00} & \underline{0.00} \\
Power & 42.86 & 37.97 & 28.57 & 57.46 & 57.14 & 60.85 & 42.86 & 55.25 & 42.86 & 56.75 & 57.14 & 45.75 & 42.86 & 61.50 & 42.86 & 62.10 \\
Rituals & 70.10 & 65.35 & 58.25 & 69.71 & 73.20 & 66.95 & 66.49 & 71.84 & 67.01 & 66.17 & 68.04 & 53.80 & 58.76 & 68.80 & 68.04 & 67.74 \\
\textbf{Stereotype} & \textbf{100.00} & \textbf{100.00} & \textbf{100.00} & \textbf{100.00} & \textbf{100.00} & \textbf{100.00} & \textbf{100.00} & \textbf{100.00} & \textbf{100.00} & \textbf{100.00} & \textbf{100.00} & \textbf{66.67} & \textbf{100.00} & \textbf{100.00} & \textbf{100.00} & \textbf{100.00} \\
Structure & 25.00 & 31.02 & 31.25 & 45.24 & 45.31 & 46.68 & 25.00 & 36.77 & 34.38 & 45.12 & 40.62 & 35.45 & 31.25 & 44.24 & 37.50 & 42.60 \\
Styles of Discourse & 50.00 & 66.83 & 12.50 & 70.16 & 37.50 & 74.64 & 37.50 & 66.96 & 25.00 & 70.71 & 25.00 & 63.42 & 12.50 & 56.71 & 37.50 & 69.77 \\
Thinking & 45.83 & 48.08 & 50.00 & 57.83 & 62.50 & 59.56 & 50.00 & 58.93 & 45.83 & 53.24 & 58.33 & 51.07 & 54.17 & 60.37 & 54.17 & 55.82 \\
Time & 33.33 & 35.84 & 0.00 & 20.03 & 33.33 & 32.88 & 33.33 & 43.99 & 33.33 & 46.95 & 33.33 & 38.43 & 33.33 & 34.32 & 33.33 & 32.88 \\
Visible Behavior & 25.00 & 48.84 & 25.00 & 46.89 & 41.67 & 62.75 & 25.00 & 44.64 & 25.00 & 46.40 & 16.67 & 36.75 & 16.67 & 42.39 & 16.67 & 45.24 \\
Ways of Dress & 66.67 & 72.36 & 25.00 & 40.91 & 75.00 & 67.62 & 33.33 & 43.87 & 58.33 & 56.04 & 58.33 & 56.37 & 16.67 & 30.79 & 33.33 & 44.14 \\
Workplace Customs & 85.71 & 77.62 & 85.71 & 90.24 & 85.71 & 75.95 & 71.43 & 84.80 & 85.71 & 82.38 & 71.43 & 67.62 & 71.43 & 89.29 & 85.71 & 80.71 \\
\textbf{Workplace Norms} & \textbf{94.44} & \textbf{88.50} & \textbf{94.44} & \textbf{87.04} & \textbf{94.44} & \textbf{88.89} & \textbf{83.33} & \textbf{81.48} & \textbf{94.44} & \textbf{96.30} & \textbf{72.22} & \textbf{69.05} & \textbf{83.33} & \textbf{85.19} & \textbf{94.44} & \textbf{88.89} \\
\bottomrule
\end{tabular}
\end{adjustbox}
\caption{Per-category (25) performance for HI-CSI (HI) and SI-CSI (SI) across Hall's cultural elements. Model abbreviations: DS-Distill = DeepSeek-R1-Distill-Qwen-32B; Claude = Claude-3.7-Sonnet; DS-R1 = DeepSeek-R1; Gemini = Gemini-2.0-Flash; OLMo-2 = OLMo-2-32B; Qwen-2.5 = Qwen-2.5-Max. Values are percentages; higher is better.}
\label{tab:hall_category_csi_iden}
\end{table*}


\begin{figure*}[ht]
  \includegraphics[width=\textwidth]{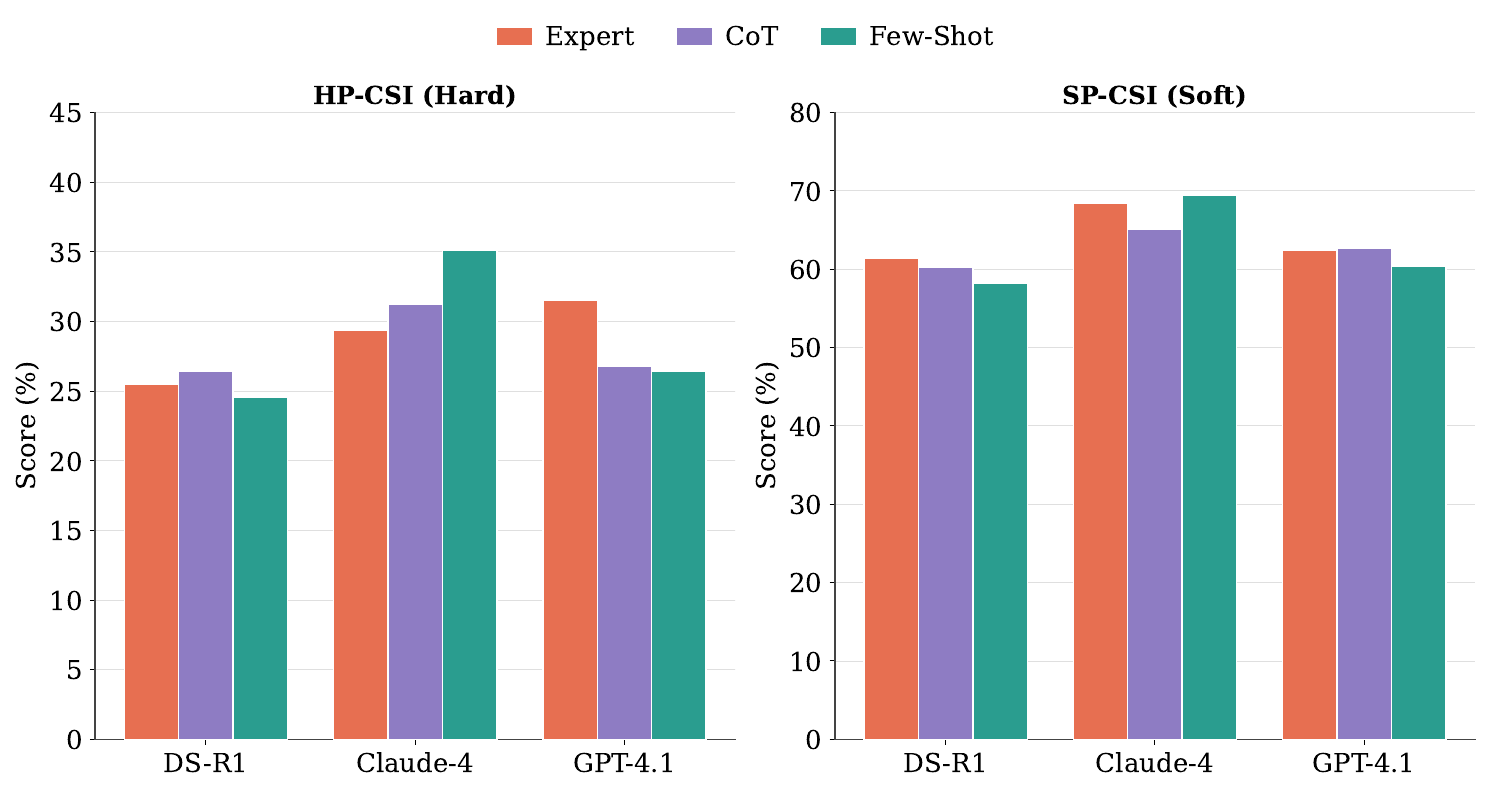}
\caption{Prompt-sensitivity ablation on the CSI Prediction task across three prompting strategies (Expert, Chain-of-Thought, Few-Shot with 3 examples) for DeepSeek-R1 (DS-R1), Claude-4-Sonnet, and GPT-4.1. Hard (HP-CSI) and soft (SP-CSI) metrics are reported. A Friedman test detects no significant difference across prompts on either metric (HP-CSI: $\chi^2(2) = 0.67$, $p = 0.72$; SP-CSI: $\chi^2(2) = 0.67$, $p = 0.72$; $n=3$ models). The absence of a consistent winner suggests that cultural reasoning gaps observed in the main results are not readily resolved by prompt engineering alone.}
\label{fig:prompt_sensitivity}
\end{figure*}


\begin{table}[ht]
\centering
\small
\begin{tabular}{@{}lcc@{}}
\toprule
\textbf{Category Pair} & \textbf{Identification} & \textbf{Prediction} \\
\midrule
Cultural Reference vs.\ Cultural Stereotype   & $0.006^{**}$ & $0.513$ \\
Cultural Reference vs.\ Cultural Value        & $0.969$ & $1.000$ \\
Cultural Reference vs.\ Social Etiquette      & $0.436$ & $0.143$ \\
Cultural Reference vs.\ Social Tradition      & $0.910$ & $1.000$ \\
Cultural Reference vs.\ Sports Tradition      & $0.436$ & $<0.001^{***}$ \\
Cultural Reference vs.\ Workplace Culture     & $0.513$ & $0.014^{*}$ \\
Cultural Stereotype vs.\ Cultural Value       & $<0.001^{***}$ & $0.742$ \\
Cultural Stereotype vs.\ Social Etiquette     & $<0.001^{***}$ & $0.993$ \\
Cultural Stereotype vs.\ Social Tradition     & $0.186$ & $0.669$ \\
Cultural Stereotype vs.\ Sports Tradition     & $0.669$ & $0.080$ \\
Cultural Stereotype vs.\ Workplace Culture    & $0.592$ & $0.742$ \\
Cultural Value vs.\ Social Etiquette          & $0.944$ & $0.296$ \\
Cultural Value vs.\ Social Tradition          & $0.363$ & $1.000$ \\
Cultural Value vs.\ Sports Tradition          & $0.058$ & $<0.001^{***}$ \\
Cultural Value vs.\ Workplace Culture         & $0.080$ & $0.042^{*}$ \\
Social Etiquette vs.\ Social Tradition        & $0.030^{*}$ & $0.237$ \\
Social Etiquette vs.\ Sports Tradition        & $0.002^{**}$ & $0.363$ \\
Social Etiquette vs.\ Workplace Culture       & $0.003^{**}$ & $0.984$ \\
Social Tradition vs.\ Sports Tradition        & $0.984$ & $<0.001^{***}$ \\
Social Tradition vs.\ Workplace Culture       & $0.993$ & $0.030^{*}$ \\
Sports Tradition vs.\ Workplace Culture       & $1.000$ & $0.864$ \\
\bottomrule
\end{tabular}
\caption{Post-hoc Nemenyi pairwise $p$-values for the CSI Identification and Prediction tasks across CSI categories. Significance levels: $^{*}p<0.05$, $^{**}p<0.01$, $^{***}p<0.001$. The contrasting patterns of significance across the two tasks support the claim that Identification and Prediction tap qualitatively different model capabilities.}
\label{tab:nemenyi_appendix}
\end{table}


\onecolumn
\subsection{CSI Adaptation}


\begin{table*}[ht]
\centering
\small
\setlength{\tabcolsep}{4.5pt}
\begin{tabular}{@{}llccc@{\hspace{8pt}}ccc@{}}
\toprule
\multirow{2}{*}{\textbf{Culture}} & \multirow{2}{*}{\textbf{Setting}} & 
\multicolumn{3}{c}{\textbf{Mean $\pm$ SE (CSI$_\mathrm{bert}$)}} & 
\multicolumn{3}{c}{\textbf{Wilcoxon $p$-value (one-sided)}} \\
\cmidrule(lr){3-5}\cmidrule(lr){6-8}
& & \textbf{V} & \textbf{SV} & \textbf{I} & \textbf{V$>$SV} & \textbf{SV$>$I} & \textbf{V$>$I} \\
\midrule
\multirow{2}{*}{Arabic} 
& Intra-lingual & $0.353 \pm 0.014$ & $0.299 \pm 0.013$ & $0.279 \pm 0.015$ 
  & $0.004^{**}$ & $0.008^{**}$ & $0.004^{**}$ \\
& Inter-lingual & $0.682 \pm 0.012$ & $0.655 \pm 0.015$ & $0.661 \pm 0.006$ 
  & $0.004^{**}$ & $0.273$ & $0.039^{*}$ \\
\midrule
\multirow{2}{*}{Chinese}
& Intra-lingual & $0.442 \pm 0.009$ & $0.346 \pm 0.010$ & $0.314 \pm 0.006$ 
  & $0.004^{**}$ & $0.008^{**}$ & $0.004^{**}$ \\
& Inter-lingual & $0.745 \pm 0.008$ & $0.717 \pm 0.005$ & $0.700 \pm 0.003$ 
  & $0.008^{**}$ & $0.008^{**}$ & $0.004^{**}$ \\
\midrule
\multirow{2}{*}{Bengali (BD)}
& Intra-lingual & $0.345 \pm 0.012$ & $0.276 \pm 0.017$ & $0.279 \pm 0.028$ 
  & $0.004^{**}$ & $0.578$ & $0.004^{**}$ \\
& Inter-lingual & $0.457 \pm 0.014$ & $0.423 \pm 0.010$ & $0.436 \pm 0.017$ 
  & $0.004^{**}$ & $0.961$ & $0.004^{**}$ \\
\midrule
\multirow{2}{*}{Bengali (WB)}
& Intra-lingual & $0.387 \pm 0.015$ & $0.305 \pm 0.019$ & $0.280 \pm 0.028$ 
  & $0.004^{**}$ & $0.074$ & $0.004^{**}$ \\
& Inter-lingual & $0.494 \pm 0.014$ & $0.441 \pm 0.012$ & $0.436 \pm 0.016$ 
  & $0.004^{**}$ & $0.422$ & $0.004^{**}$ \\
\bottomrule
\end{tabular}
\caption{CSI Adaptation performance (CSI$_\mathrm{bert}$) across Hall's cultural triad (V = Visible, SV = Semi-visible, I = Invisible) for four target cultures, in both intra- and inter-lingual settings. Mean $\pm$ standard error reported across 8 LLMs. One-sided Wilcoxon signed-rank tests assess whether higher-visibility levels yield higher scores. Significance: $^{*}p<0.05$, $^{**}p<0.01$, $^{***}p<0.001$. The Visible-vs-Invisible decline is significant in all eight (culture, setting) combinations; the Semi-visible-vs-Invisible step is less consistent, particularly for Bengali and the inter-lingual settings.}
\label{tab:hall_adaptation}
\end{table*}

\begin{figure*}[ht]
    \centering
    \begin{subfigure}{0.49\textwidth}
        \centering
        \includegraphics[width=\textwidth]{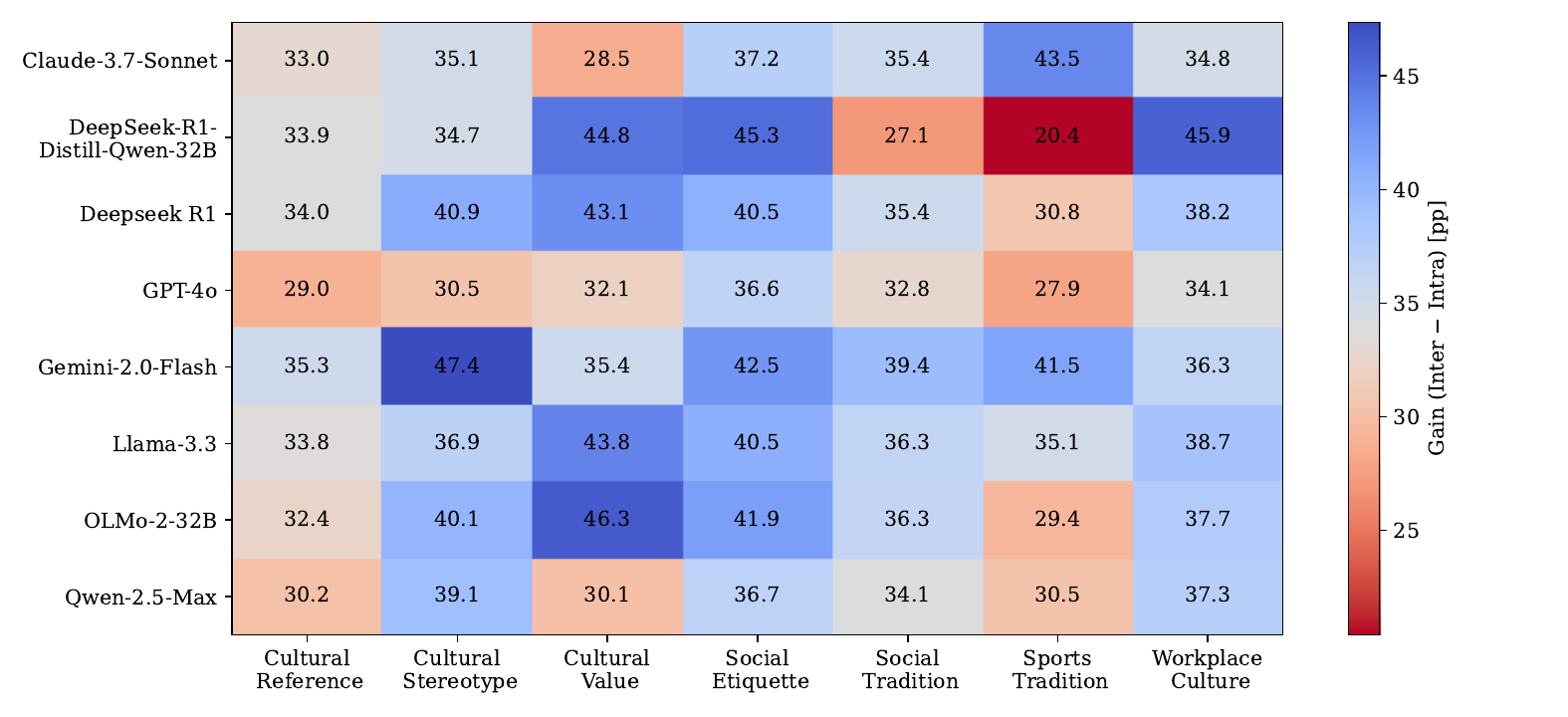}
        \caption{Inter-Intra CSI Adaptation Gain for Arabic}
        \label{fig:arabic_gain_adapt}
    \end{subfigure}
    \begin{subfigure}{0.49\textwidth}
        \centering
        \includegraphics[width=\textwidth]{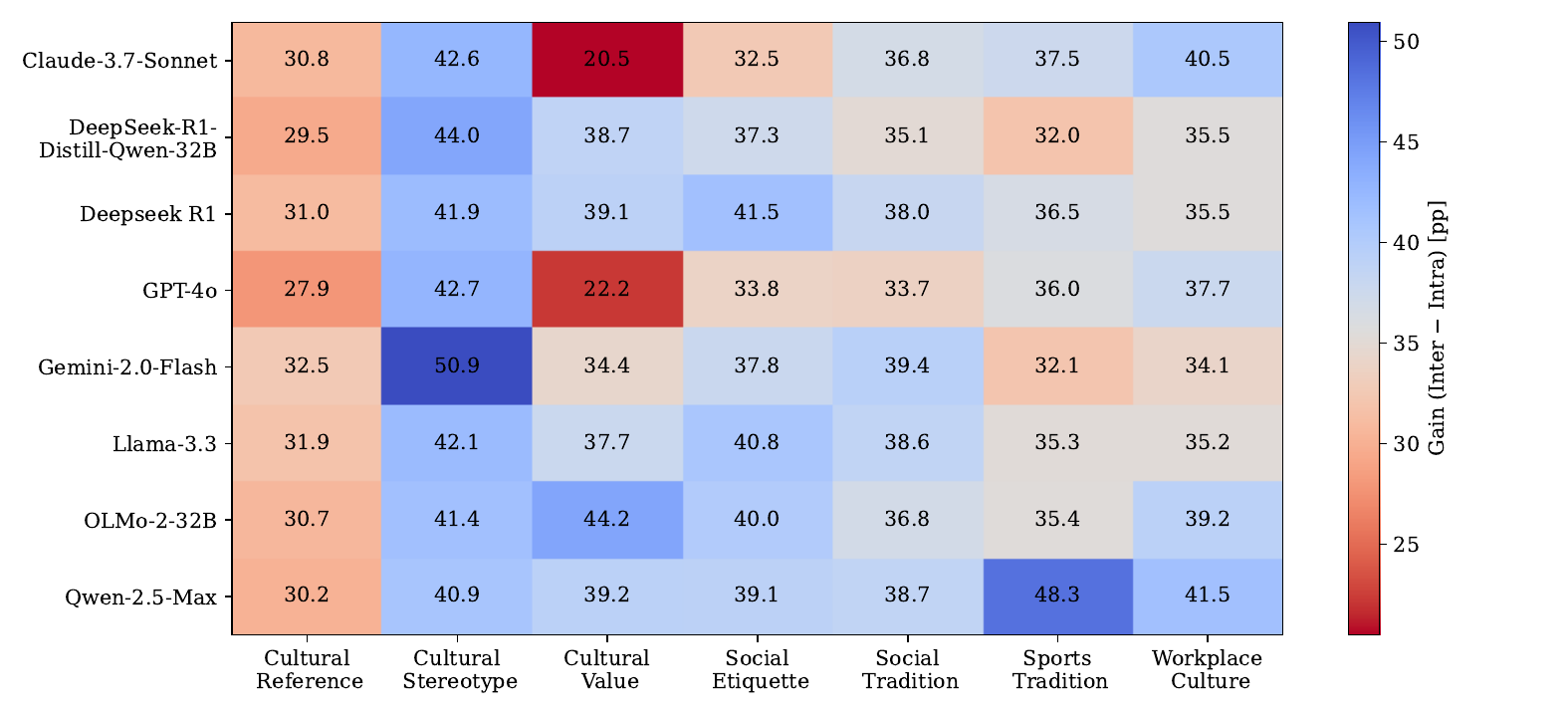}
        \caption{Inter-Intra CSI Adaptation Gain for Chinese}
        \label{fig:chinese_gain_adapt}
    \end{subfigure}

    \caption{Inter--Intra CSI adaptation gain for Arabic and Chinese. The heatmap reports the difference between inter-lingual and intra-lingual CSI adaptation scores (Inter $-$ Intra) under the CSI$_\mathrm{bert}$ metric across models and CSI categories. Across both cultures, all 56 (model, category) cells show positive gains (median gain $\approx 36$ for Arabic and $\approx 37$ for Chinese; one-sided Wilcoxon signed-rank test against zero, $p < 10^{-10}$ for each culture), indicating that LLMs adapt to Arabic and Chinese cultural content significantly more effectively in inter-lingual settings than in intra-lingual ones.}
    \label{fig:inter_vs_intra_arabic_chinese}
\end{figure*}

\begin{table*}[ht]
\centering
\small
\setlength{\tabcolsep}{6pt}
\renewcommand{\arraystretch}{1.15}
\begin{tabular}{l*4{c}:*4{c}}
\toprule
& \multicolumn{2}{c}{\textbf{Arabic}}
& \multicolumn{2}{c}{\textbf{Chinese}}
& \multicolumn{2}{c}{\textbf{Bengali (WB)}}
& \multicolumn{2}{c}{\textbf{Bengali (BD)}} \\
\cmidrule(lr){2-3}\cmidrule(lr){4-5}\cmidrule(lr){6-7}\cmidrule(lr){8-9}
Model & Intra & Inter & Intra & Inter & Intra & Inter & Intra & Inter \\
\midrule
DeepSeek-R1
& 69.02 & \underline{81.09}
& 72.46 & \underline{83.47}
& \textbf{84.40} & \textbf{74.40}
& \textbf{83.54} & \textbf{74.13} \\
Gemini-2.0-Flash
& 68.91 & 80.78
& 70.65 & 83.08
& 76.84 & 70.30
& 76.33 & 70.14 \\
GPT-4o
& \textbf{72.76} & 81.08
& \textbf{74.66} & \underline{83.81}
& 81.69 & 70.49
& 80.56 & 70.16 \\
Llama-3.3
& 68.95 & 80.90
& 72.74 & 83.59
& \underline{84.32} & \underline{74.40}
& \underline{83.52} & \underline{74.12} \\
Qwen-2.5-Max
& 70.54 & 80.51
& 67.44 & 83.75
& 64.68 & 70.60
& 64.02 & 70.42 \\
OLMo-2-32B
& 69.01 & 81.09
& 72.71 & 83.56
& 76.11 & 70.80
& 75.27 & 70.61 \\
Claude-3.7-Sonnet
& \underline{72.65} & \textbf{81.64}
& 73.88 & \textbf{83.91}
& 82.49 & 73.01
& 80.92 & 72.24 \\
D-R1-D-Qwen-32B
& 68.85 & 75.96
& \underline{74.15} & 83.42
& 78.16 & 62.27
& 77.10 & 62.14 \\
\midrule
Average
& 70.08 & 80.38
& 72.34 & 83.58
& 78.59 & 70.78
& 77.66 & 70.50 \\
\bottomrule
\end{tabular}
\caption{Sentence-level CSI adaptation performance (\%) across models measured using SENT$_\mathrm{bert}$, reported in both intra-lingual and inter-lingual settings across four target cultures. Best results per column are shown in bold, and second-best results are underlined. SENT$_\mathrm{bert}$ scores remain uniformly high across all settings because adaptation typically involves localised substitutions that leave overall sentence structure largely unchanged; CSI-level scores (CSI$_\mathrm{bert}$) in Table~\ref{tab:csi_adaptation_result} are therefore the more discriminative metric. Abbreviations: Bangladesh (\textit{BD}), West Bengal (\textit{WB}).}
\label{tab:sentbert_adaptation_result}
\end{table*}

\begin{figure*}[ht]
    \centering
    \begin{subfigure}{0.49\textwidth}
        \centering
        \includegraphics[width=\textwidth]{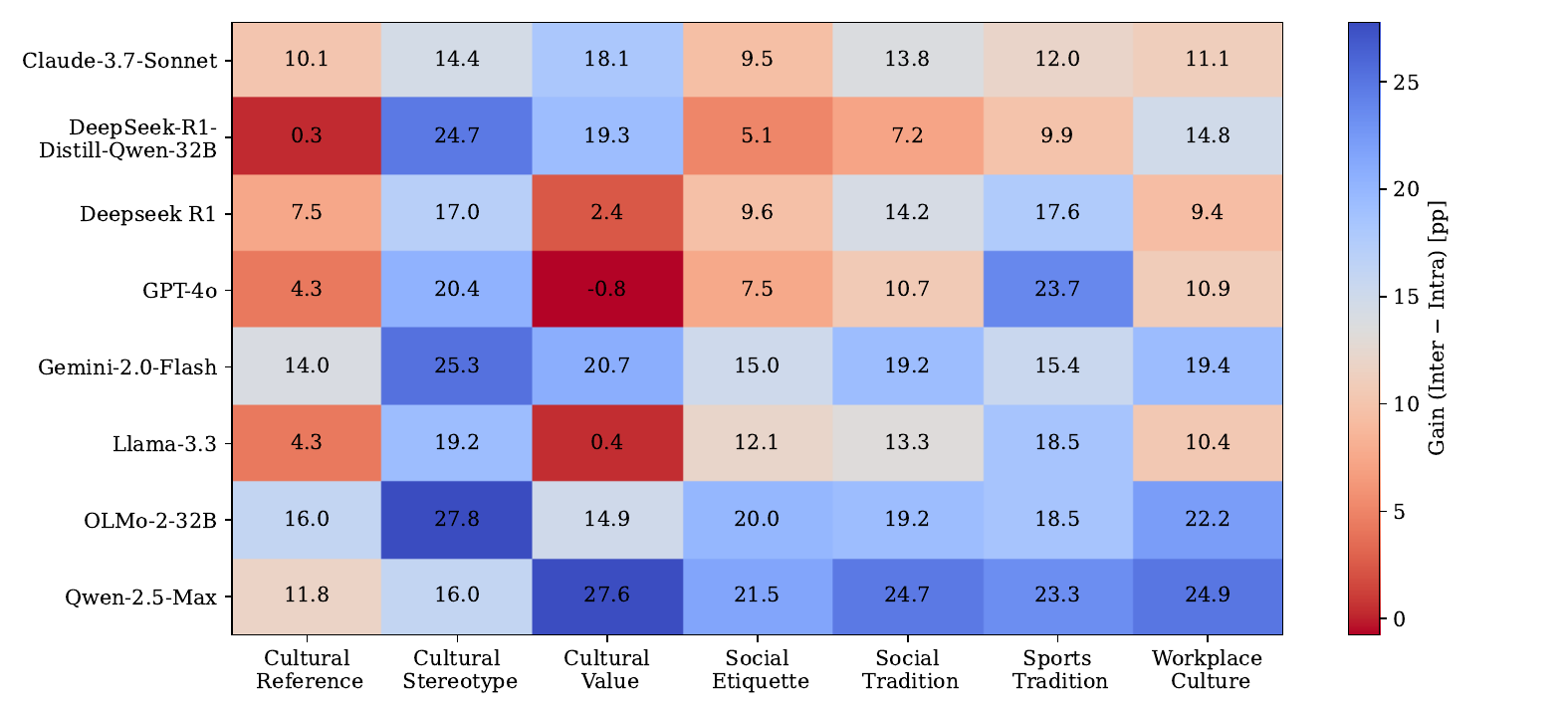}
        \caption{Inter-Intra CSI Adaptation Gain for Bn(West Bengal)}
        \label{fig:west_bengal_gain_adapt}
    \end{subfigure}
    \begin{subfigure}{0.49\textwidth}
        \centering
        \includegraphics[width=\textwidth]{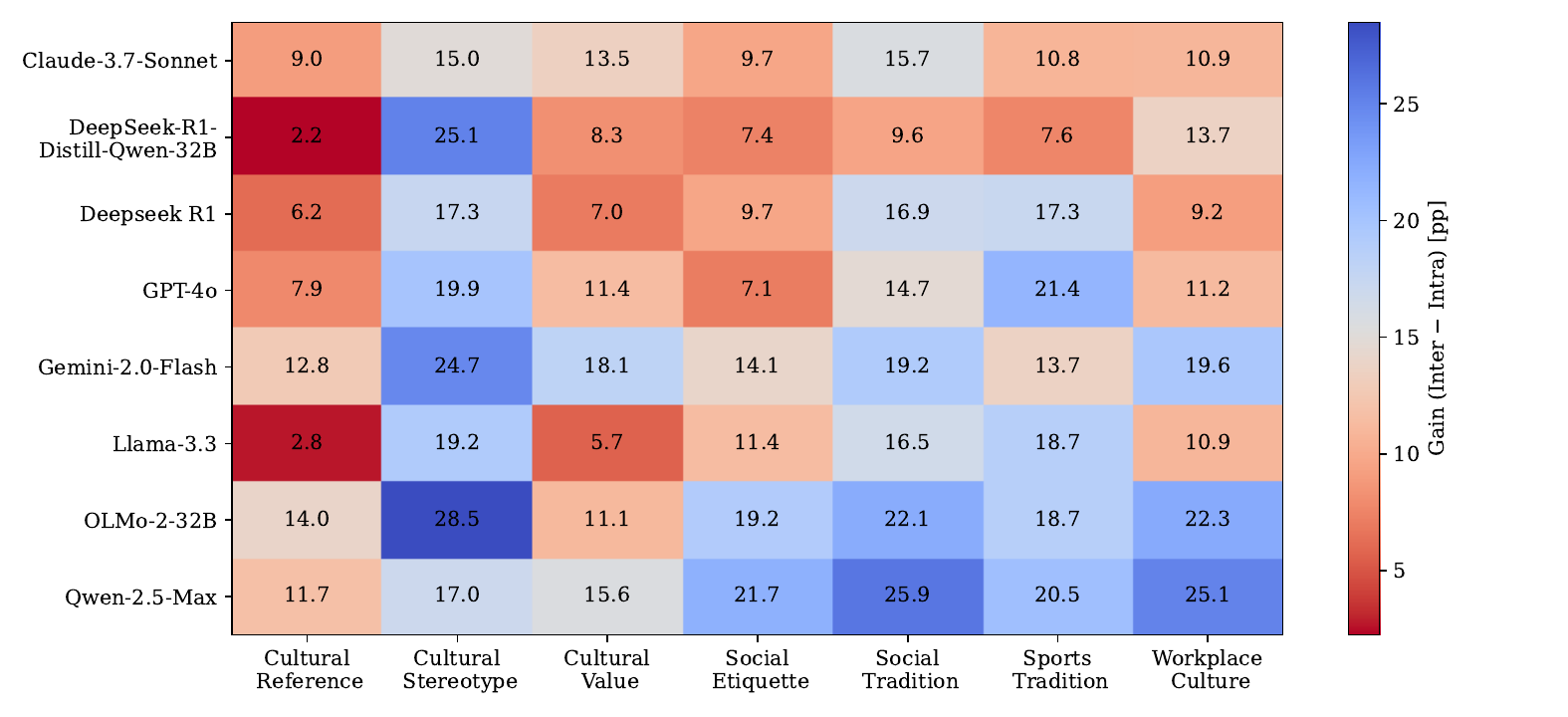}
        \caption{Inter-Intra CSI Adaptation Gain for Bn(Bangladesh)}
        \label{fig:bangladesh_gain_adapt}
    \end{subfigure}

    \caption{Inter--Intra CSI adaptation gain for Bengali (West Bengal and Bangladesh). The heatmap reports the difference between inter-lingual and intra-lingual CSI adaptation scores (Inter $-$ Intra) under the CSI$_\mathrm{bert}$ metric across models and CSI categories. Positive gains dominate across both variants (55/56 cells for WB, 56/56 for BD; median gain $\approx 14$ in both; one-sided Wilcoxon signed-rank test against zero, $p < 10^{-10}$ for each), confirming that adaptation in the target language consistently outperforms intra-lingual adaptation. Note that gain magnitudes are markedly smaller for Bengali than for Arabic or Chinese (Figure~\ref{fig:inter_vs_intra_arabic_chinese}), reflecting weaker target-language proficiency in low-resource settings.}
    \label{fig:inter_vs_intra_wb_bd}
\end{figure*}




\begin{table*}[ht]
\centering
\small
\setlength{\tabcolsep}{4.5pt}
\renewcommand{\arraystretch}{1.15}
\begin{tabularx}{\textwidth}{
l l
>{\raggedright\arraybackslash}X c
>{\raggedright\arraybackslash}X c}
\toprule
\textbf{Language} & \textbf{Model} &
\textbf{Best Method} & \textbf{Score} &
\textbf{Worst Method} & \textbf{Score} \\
\midrule

Arabic & DeepSeek-R1
& Couplet (Cultural Equivalent + Paraphrase)
& 1.00
& Cultural Equivalent
& 0.50 \\

Arabic & Gemini-2.0-Flash
& Couplet (Cultural Equivalent + Classifier)
& 0.86
& Couplet (Cultural Equivalent + Neutralization)
& 0.57 \\

Arabic & GPT-4o
& Cultural Equivalent
& 0.79
& Cultural Equivalent + Componential Analysis
& 0.53 \\

Arabic & Llama-3.3
& Couplet (Cultural Equivalent + Accepted Standard Translation)
& 1.00
& Couplet (Cultural Equivalent + Neutralization)
& 0.50 \\

Arabic & OLMo-2-32B
& Couplet (Cultural Equivalent + Literal Translation)
& 1.00
& Cultural Equivalent + Labeling
& 0.50 \\

Arabic & Qwen-2.5-Max
& Transference
& 0.78
& Paraphrase
& 0.57 \\

\midrule
Bengali & Claude-3.7-Sonnet
& Couplet (Transference + Naturalization)
& 1.00
& Componential Analysis
& 0.46 \\

Bengali & DeepSeek-R1
& Couplet (Cultural Equivalent + Accepted Standard Translation)
& 0.90
& Cultural Equivalent
& 0.41 \\

Bengali & Gemini-2.0-Flash
& Cultural Equivalent + Labeling
& 0.80
& Cultural Equivalent + Label
& 0.43 \\

Bengali & GPT-4o
& Literal Translation
& 0.88
& Paraphrase
& 0.41 \\

Bengali & Llama-3.3
& Couplet (Cultural Equivalent + Literal Translation)
& 0.90
& Cultural Equivalent
& 0.43 \\

Bengali & Qwen-2.5-Max
& Couplet (Cultural Equivalent + Paraphrase)
& 0.81
& Transference
& 0.42 \\

\midrule
Chinese & DeepSeek-R1
& Cultural Equivalent
& 1.00
& Cultural Equivalent
& 0.50 \\

Chinese & Gemini-2.0-Flash
& Couplet (Transference + Labeling)
& 0.79
& Couplet (Cultural Equivalent + Paraphrase)
& 0.53 \\

Chinese & GPT-4o
& Couplet (Cultural Equivalent + Accepted Standard Translation)
& 1.00
& Couplet (Cultural Equivalent + Paraphrase)
& 0.50 \\

Chinese & Llama-3.3
& Cultural Equivalent
& 1.00
& Cultural Equivalent
& 0.50 \\

Chinese & OLMo-2-32B
& Cultural Equivalent
& 1.00
& Cultural Equivalent
& 0.51 \\

Chinese & Qwen-2.5-Max
& Couplet (Cultural Equivalent + Naturalization)
& 1.00
& Couplet (Cultural Equivalent + Literal Translation)
& 0.51 \\

\bottomrule
\end{tabularx}
\caption{Best- and worst-performing adaptation methods (by average CSI score) for each language–model pair. ``Couplet'' denotes a combination of two translation strategies. Cultural Equivalent is abbreviated as CE in the main text.}
\label{tab:best_worst_adaptation_methods}
\end{table*}

\onecolumn
\section{Prompts Used for Evaluation}
\label{sec:prompts_appendix}

\subsection{CSI Identification}

\begin{figure*}[ht]
\small
\centering
\begin{tcolorbox}[colback=blue!10!white, colframe=blue!5!black, title=]
You are a highly skilled and culturally aware agent with expertise in identifying Culture-Specific Items (CSIs). CSIs are words, phrases, or concepts that are specific to a particular culture or carry significant cultural meaning. They depend on cultural context and can vary widely across cultures. CSIs encompass natural elements like animals, plants, and environmental features (ecology); physical objects like food, clothing, housing, and tools (material culture); activities related to work and leisure (social culture); political, social, legal, religious, and artistic concepts (organizations, customs, and ideas); and non-verbal communication or behaviors like gestures and habits.\\

Your task is to identify the CSI(s) in the given sentence and enclose them with `<CSI>` and `</CSI>` tags. The CSIs in this task will mostly pertain to American/Western/British Culture, so prioritize identifying items specific to these cultural contexts. Do not add any explanations, notes, or unnecessary text—only provide the revised sentence with the CSI(s) properly tagged.\\ 

Now process the following sentences exactly as instructed:
Input: 

\end{tcolorbox}
\caption{Evaluation Prompt for CSI Identification Task.}    
\label{fig:csi_iden_eval_prompt}
\end{figure*}

\subsection{CSI Prediction}

\begin{figure*}[h]
\small
\centering
\begin{tcolorbox}[colback=blue!10!white, colframe=blue!5!black, title=]
You are a highly skilled and culturally aware agent with expertise in predicting Culture-Specific Items (CSIs). CSIs are words, phrases, or concepts that are specific to a particular culture or carry significant cultural meaning. They depend on cultural context and can vary widely across cultures. CSIs encompass natural elements like animals, plants, and environmental features (ecology); physical objects like food, clothing, housing, and tools (material culture); activities related to work and leisure (social culture); political, social, legal, religious, and artistic concepts (organizations, customs, and ideas); and non-verbal communication or behaviors like gestures and habits.\\

You’ll be given inputs where each sentence will contain a `[MASK]` tag, where the CSI has been masked. The CSI is surrounded with a beginning <CSI> tag and an ending </CSI> tag. Your task is to predict the most appropriate CSI to fill in the `[MASK]` based on the surrounding context. The CSIs in this task will mostly pertain to American/Western/British Culture, so prioritize predicting items specific to these cultural contexts.\\

Replace the [MASK] with the predicted CSI. Do not add any explanations, notes, or unnecessary text—only provide the revised sentence with the predicted CSI filled in.

Now process the following sentences exactly as instructed:

Input:

\end{tcolorbox}
\caption{Evaluation Prompt for CSI Prediction Task.}    
\label{fig:csi_pred_eval_prompt}
\end{figure*}

\onecolumn
\subsection{CSI Adaptation}

\begin{figure*}[ht]
\small
\centering
\begin{tcolorbox}[colback=blue!10!white, colframe=blue!5!black, title=]
You are an expert in cultural adaptation and linguistic transformation with a deep understanding of Bengali/Chinese/Arabic culture, traditions, and social practices. You specialize in providing culturally sensitive adaptations that align with the values, norms, and etiquette of Bengali/Chinese/Arabic people. You will be given an input sentence containing Culture-Specific Items (CSIs) that mostly pertain to American/Western/British Culture.  Your task is to adapt sentences containing <CSI> tags into the context of Bengali/Chinese/Arabic society, ensuring both cultural relevance and linguistic accuracy. You can use one or a combination of the following methods appropriate for adaptation:  
\begin{enumerate}
    \item  Transference: Keep the original word unchanged in the adaptation.
    \item Cultural Equivalent: Use a similar term from the target culture.
    \item Neutralization: Explain what the term means or does.
    \item Literal Translation: Translating the word directly to target culture.
    \item Label: Add a brief explanation or tag to the term.
    \item Naturalization: Adapt the word to fit the target language's spelling or sound.
    \item Componential Analysis: Break the term into parts and explain each.
    \item Deletion: Remove unnecessary words or phrases.
    \item Couplet: Combine two adaptation methods
    \item Accepted Standard Translation: Use a commonly accepted adaptation.
    \item Paraphrase or Gloss: Give a longer explanation or footnote.
    \item Classifier: Add a general category to clarify the term’s meaning.
\end{enumerate}
Your task is to adapt the following sentences containing <CSI> tags into Bengali/Chinese/Arabic culture. Replace inside the <CSI> tags with culturally relevant practices, behaviors, or terms. Do not remove the <CSI> tags.  In case there is no cultural equivalent in the target culture, return `Non-transferable'.\\

Provide both Intra-lingual adaptation (Contextualized for Bengali culture but written in English) and Inter-lingual adaptation (Translated and adapted into Bengali/Chinese/Arabic for each input. 
Ensure cultural accuracy, politeness, and respect in all adaptations. Also output what method you used for the adaptation. The output format should be [Intra-lingual adaptation, Inter-lingual adaptation, Adaptation method used]. Do not add any explanation. 

Now adapt the following sentences:

Input:

\end{tcolorbox}
\caption{Evaluation Prompt for CSI Adaptation Task.}    
\label{fig:csi_adapt_eval_prompt}
\end{figure*}

\end{document}